\documentclass[letterpaper]{article} 
\usepackage[submission]{aaai23}  
\usepackage{times}  
\usepackage{helvet}  
\usepackage{courier}  
\usepackage[hyphens]{url}  
\usepackage{graphicx} 
\usepackage{natbib}  
\usepackage{caption} 
\usepackage{algorithm}
\usepackage{algorithmic}

\usepackage{graphicx}
\usepackage{booktabs}
\usepackage{multirow}
\usepackage{mathtools}

\usepackage{pifont}
\usepackage{pdfrender}
\newcommand{\cmark}{\ding{51}}
\newcommand{\xmark}{\ding{55}}%
\usepackage{colortbl}

\usepackage{cite}

\newcommand{\stdv}[1]{\scriptsize$\pm$#1}

\newcommand*{\boldcheckmark}{%
  \textpdfrender{
    TextRenderingMode=FillStroke,
    LineWidth=.5pt, 
  }{\cmark}%
}
\newcommand*{\boldxmark}{%
  \textpdfrender{
    TextRenderingMode=FillStroke,
    LineWidth=.5pt, 
  }{\xmark}%
}

\definecolor{Gray}{gray}{0.95}

\definecolor{MyGreen}{rgb}{0,0.6,0.3}
\definecolor{blackpink}{rgb}{0.6,0,0.6}
\definecolor{MyRed}{RGB}{200, 0, 0}

\usepackage{newfloat}
\usepackage{listings}
\DeclareCaptionStyle{ruled}{labelfont=normalfont,labelsep=colon,strut=off} 
\floatstyle{ruled}
\newfloat{listing}{tb}{lst}{}
\floatname{listing}{Listing}
\title{Revisiting the Importance of Amplifying Bias for Debiasing}

\author{
    Jungsoo Lee\equalcontrib\textsuperscript{\rm 1,2},
    Jeonghoon Park\equalcontrib\textsuperscript{\rm 1,2},
    Daeyoung Kim\equalcontrib\textsuperscript{\rm 1}, \\
    Juyoung Lee\textsuperscript{\rm 2},
    Edward Choi\textsuperscript{\rm 1},
    Jaegul Choo\textsuperscript{\rm 1}
}
\affiliations{
    \textsuperscript{\rm 1}
    KAIST 
    \textsuperscript{\rm 2}
    Kakao Enterprise, South Korea \\
    \textsuperscript{\rm 1}
    \{bebeto, jeonghoon\texttt{\_}park, daeyoung\texttt{.}k, edwardchoi, jchoo\}@kaist.ac.kr \\
    \textsuperscript{\rm 2}
    \{michael.jy\}@kakaoenterprise.com
}

\usepackage{bibentry}

\begin{document}

\maketitle

\begin{abstract}
In image classification, \textit{debiasing} aims to train a classifier to be less susceptible to dataset bias, the strong correlation between peripheral attributes of data samples and a target class. For example, even if the frog class in the dataset mainly consists of frog images with a swamp background (\textit{i.e.,} bias-aligned samples), a debiased classifier should be able to correctly classify a frog at a beach (\textit{i.e.,} bias-conflicting samples). Recent debiasing approaches commonly use two components for debiasing, a biased model $f_B$ and a debiased model $f_D$. $f_B$ is trained to focus on bias-aligned samples (\textit{i.e.,} overfitted to the bias) while $f_D$ is mainly trained with bias-conflicting samples by concentrating on samples which $f_B$ fails to learn, leading $f_D$ to be less susceptible to the dataset bias. While the state-of-the-art debiasing techniques have aimed to better train $f_D$, we focus on training $f_B$, an overlooked component until now. Our empirical analysis reveals that removing the bias-conflicting samples from the training set for $f_B$ is important for improving the debiasing performance of $f_D$. 
This is due to the fact that the bias-conflicting samples work as noisy samples for amplifying the bias for $f_B$ since those samples do not include the bias attribute.
To this end, we propose a \emph{simple yet effective} data sample selection method which removes the bias-conflicting samples to construct a bias-amplified dataset for training $f_B$. 
Our data sample selection method can be directly applied to existing reweighting-based debiasing approaches, obtaining consistent performance boost and achieving the state-of-the-art performance on both synthetic and real-world datasets. 
\end{abstract}
\vspace{-0.5cm}

\section{Introduction}
\label{sec:introduction}
When there exists a correlation between peripheral attributes and labels which is referred to as \textit{dataset bias}~\cite{unbiaslook2011torralba} in the training dataset, image classification models often heavily rely on such a bias.
Dataset bias is caused when the majority of data samples include bias attributes, the visual attributes that frequently co-occur with the target class but not innately defining it~\cite{disentangled}.
For example, frogs are commonly observed in swamps (bias attribute), but frogs can also be found in other places such as grasses or beaches. 
In such a case, the image classification model trained with the biased dataset could use swamps as the visual cue for classifying frogs. 
In other words, it may fail to correctly classify frog images in other places. 
To mitigate such an issue, debiasing aims to train the image classification model to learn the intrinsic attributes, the visual attributes which inherently define a target class, such as the legs or eyes of frogs. 

\begin{figure}[t!]
    \centering
    \includegraphics[width=1.0\columnwidth]{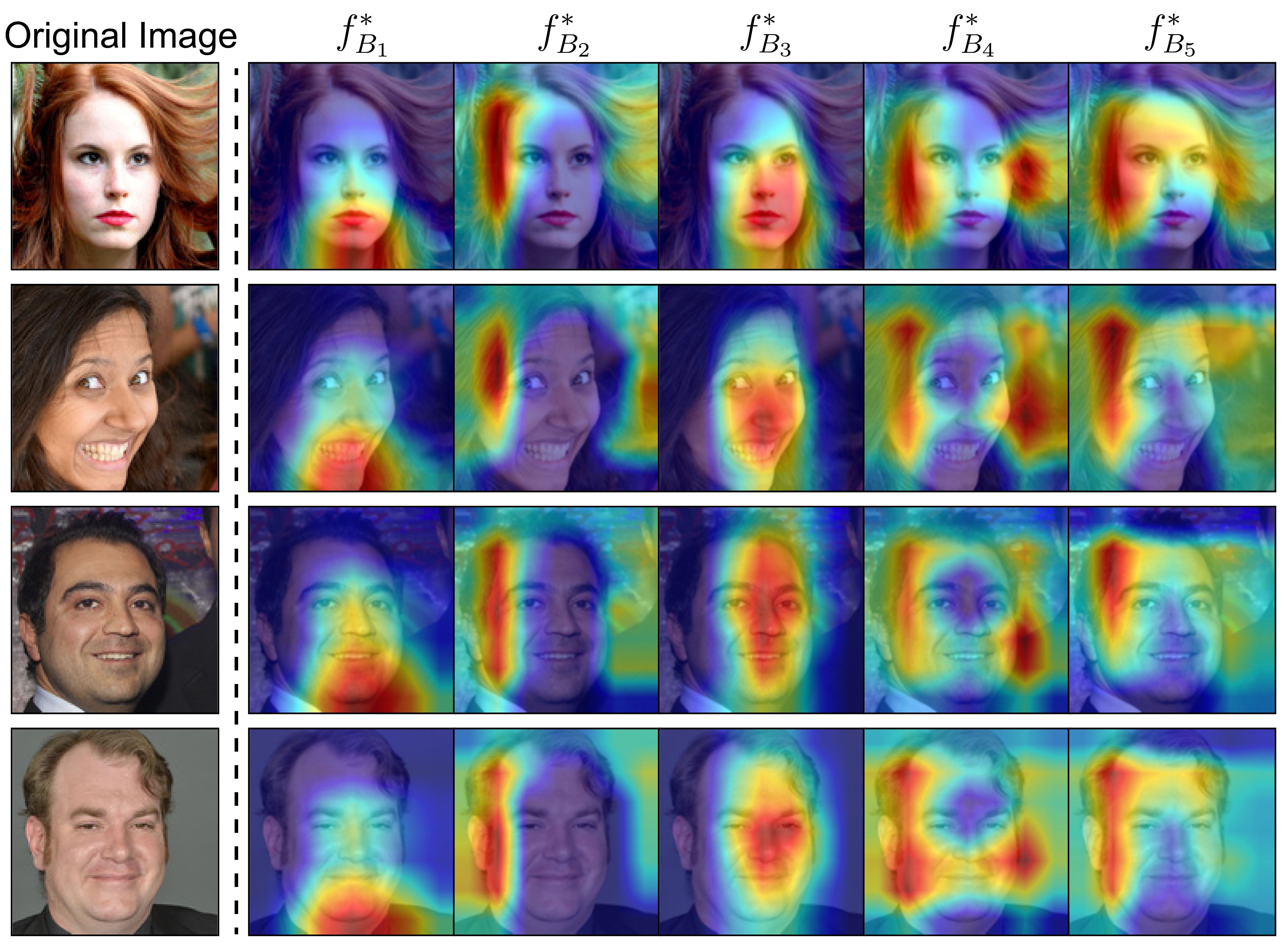}
    \vspace{-0.4cm}
    \caption{Visualization results of GradCAM applied to biased models with different initialization~(\textit{i.e.,} $f_{B_1}^*,\dots,f_{B_5}^*$). The first column indicates the original images. Starting from the second column, we observe that each biased model with different initialization focuses on different visual attributes.}
    \label{fig:gradcam}
    \vspace{-0.6cm}
\end{figure}

In a biased dataset, the data samples without the bias attribute (\textit{i.e.,} bias-conflicting samples) such as frogs on grasses or beaches are excessively scarce compared to the samples including the bias attributes (\textit{i.e.,} bias-aligned samples).
Due to the scarcity, existing state-of-the-art debiasing studies~\cite{nam2020learning, disentangled} train a given model by \emph{reweighting} the data samples which refers to imposing 1) high weight on losses of the bias-conflicting samples and 2) low weight on those of the bias-aligned ones.
For example, Nam~\textit{et al.}~\shortcite{nam2020learning} reweight the data samples based on the finding that the bias attributes are \emph{easy to learn} compared to the intrinsic attributes.
To be more specific, they intentionally train a biased model $f_B$ to be overfitted to the easily learned bias attribute.
Then, they utilize $f_B$ for computing a reweighting value $w$ for each training sample, which the value is designed to be high for samples $f_B$ fails to classify (\textit{i.e.,} bias-conflicting samples).
The data items are reweighted with $w$ during training the model $f_D$ to learn the debiased representation.
In this regard, how well $f_B$ is overfitted to the bias attribute influences the debiasing performance of $f_D$ since it determines the reweighting value.

However, our careful analysis points out that $f_B$ used in the existing reweighting-based approaches~\cite{nam2020learning, disentangled} fail to maximally exploit the bias attribute. 
They utilize a loss function which is designed to emphasize the bias-aligned samples in order to overfit $f_B$ to bias attributes.
Despite such a design, even the small number of bias-conflicting samples still interfere with $f_B$ from being overfitted to the bias attribute since they work as noisy samples for learning the bias.
In spite of the importance of such an issue, none of the previous studies shed light on removing the bias-conflicting samples from training sets for overfitting $f_B$ to the bias attribute to improve the debiasing performance, especially challenging without explicit bias labels (\textit{i.e.,} annotations of bias attributes) or prior knowledge on certain bias.

To this end, we propose a \emph{simple yet effective} biased sample selection method that builds a refined dataset which discards bias-conflicting samples and mainly includes the bias-aligned ones in order to amplify bias when training $f_B$. 
While the bias attribute (\textit{e.g.,} gender) is easy to learn~\cite{nam2020learning}, it is composed of multiple visual attributes (\textit{e.g.,} make-up, hairstyle, beards).
We found that it is challenging for a biased model to consider multiple visual attributes comprehensively for making biased predictions.
To be more specific, as shown in Fig.~\ref{fig:gradcam}, differently initialized biased models only utilize certain visual attributes for making biased predictions.
Additionally, these attributes are different among the models (\textit{e.g.,} one focuses on the lips mainly while the other concentrates on the hairstyle for predicting the gender).
Such a finding aligns with the previous studies which found that deep neural networks learn in different ways with different random initialization~\cite{ensemble-help-generalization, ensemble-loss-landscape, ensemble-search-uncertainty}.
This observation indicates that we can better understand the bias by capturing diverse visual attributes of the bias by utilizing the predictions of multiple biased models.

The procedure of our proposed method is as follows.
First, in order to capture diverse visual attributes of a bias, we pretrain multiple biased models with different random initialization for a small number of iterations by using the \emph{easy-to-learn} property of bias attributes~\cite{nam2020learning}.
By utilizing the predictions of the differently initialized biased models, we refine the train dataset with the bias-conflicting samples discarded.  
The newly refined dataset which mainly includes bias-aligned samples is then used to train $f_B$.
Training with the bias-amplified dataset encourages $f_B$ to maximally exploit the bias attribute when making predictions, and improve the debiasing performance of $f_D$ overall.

In summary, the main contributions of our paper are as follows:
\begin{itemize}
    \item Based on our preliminary analysis, we reveal that how well $f_B$ is overfitted to the bias influences the debiasing performance crucially, an important observation overlooked in the previous reweighting-based approaches.
    \item We propose a \emph{simple yet effective} biased sample selection method which better captures a bias attribute by considering multiple visual attributes of a bias.
    \item Our method can be easily adopted to existing reweighting-based approaches, and we achieve the new state-of-the-art performances on both synthetic and real-world datasets.  
\end{itemize}

\section{Related Work}

Existing early studies of debiasing explicitly use bias labels during training~\cite{LNL, EnD, sagawa2019distributionally} or implicitly predefine the bias types (\textit{e.g.,} focusing on mitigating the color bias)~\cite{wang2018hex, geirhos2018imagenettrained, bahng2019rebias}. 
Bias labels or prior knowledge on the bias types are generally used to identify bias-conflicting samples. 
Although not utilizing explicit bias labels, ReBias~\cite{bahng2019rebias} predefines a certain bias type (\textit{e.g.,} color and texture) and focuses on mitigating such bias by leveraging a color- and texture-oriented network with small receptive fields~\cite{brendel2018bagnets} to capture the predefined color or texture bias.
However, acquiring bias labels or predefining a bias type 1) necessitates humans to identify the bias type of a given dataset and 2) limits the debiasing performance on unknown bias types~\cite{disentangled}. 

Recent debiasing studies proposed several methods to address such an issue~\cite{darlow2020latent, huangRSC2020, nam2020learning, disentangled}. Nam~\textit{et al.}~\shortcite{nam2020learning} propose LfF which identifies the bias-conflicting samples based on the intuitive finding that the bias attributes are \emph{easy to learn} compared to the intrinsic attributes.
By using the fact that $f_B$ outputs a relatively high loss value for bias-conflicting samples, they impose high weight on (\textit{i.e.,} emphasize) bias-conflicting samples and low weight on the bias-aligned samples during training $f_D$. 
Lee~\textit{et al.}~\shortcite{disentangled} augment the bias-conflicting samples via disentangled feature-level augmentation, emphasizing them along with the bias-conflicting samples in the original training set by using the reweighting method.
Although the previous studies utilize $f_B$ for computing the reweighting value, we reveal that they overlooked the importance of amplifying bias for $f_B$, 
crucial for improving the debiasing performance of $f_D$.

\section{Importance of Amplifying Bias}
\label{sec:motivation}

\begin{figure*}[t]
    \centering
    \includegraphics[width=0.85\textwidth]{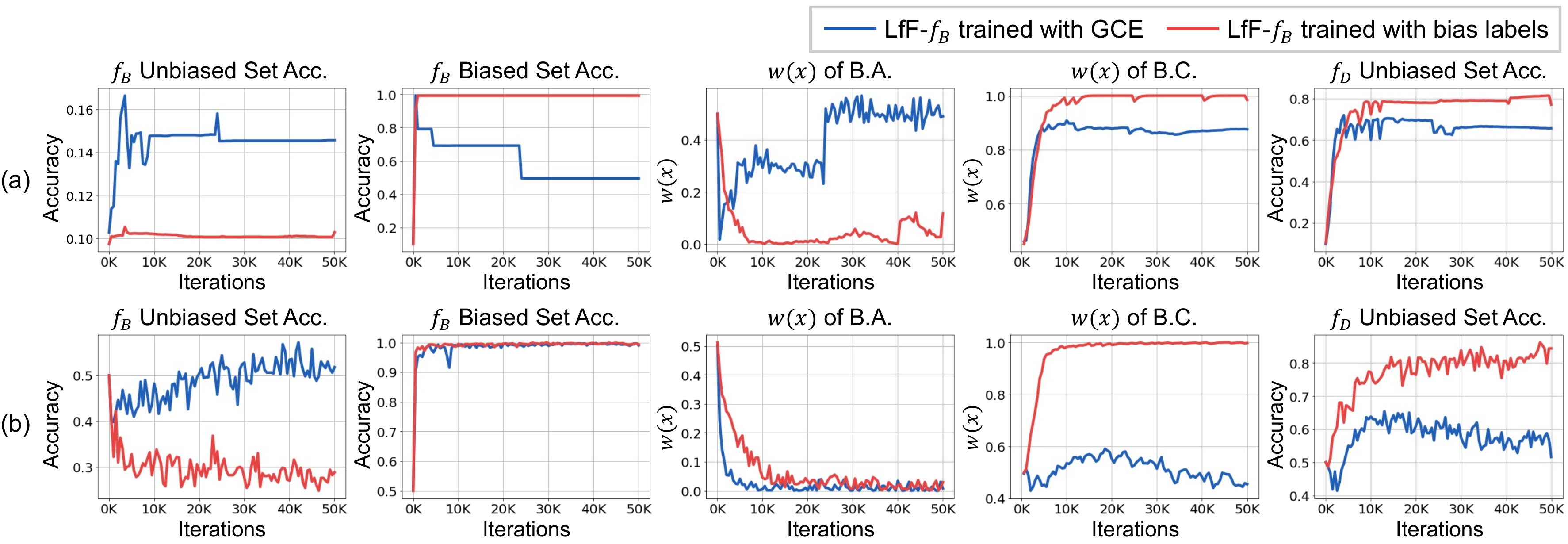}
    \vspace{-0.15cm}    
    \caption{Comparison of LfF utilizing $f_B$ trained with 1) GCE loss (blue) and 2) explicit bias labels (red). (a) and (b) indicate the results on Colored MNIST and BFFHQ, respectively. Starting from the first column, each graph represents the 1) unbiased test set accuracy of $f_B$, 2) biased test set accuracy of $f_B$, 3) averaged reweighting value $w(x)$ of bias-aligned samples (abbreviated as B.A.), 4) that of bias-conflicting samples (abbreviated as B.C.), and 5) unbiased test set accuracy of $f_D$.}
    \vspace{-0.5cm}
    \label{fig:reweighting-motivation}
\end{figure*}

\subsection{Background}
\label{sec:background-of-reweighting}
\noindent \textbf{Overfitting model to the bias. \enskip}
Since annotating bias labels or identifying the bias types in advance is challenging and labor intensive~\cite{disentangled}, recent studies leverage the Generalized Cross Entropy (GCE) loss~\cite{zhang2018generalized} that does not require such information for amplifying the bias~\cite{nam2020learning, disentangled}.
The GCE loss is defined as: 
\begin{equation}
    \mathcal{L_\text{GCE}}(p(x;\theta), y) = \frac{1-p_y(x;\theta)^{q}}{q}, 
\end{equation} 
where $q$ is a scalar value which controls the degree of amplification, and $p(x; \theta)$ and $p_{y}(x; \theta)$ are the softmax outputs of the network parameterized by $\theta$ and the softmax probability of the target class $y$, respectively.
The GCE loss assigns high weights on the gradients of the samples with the high prediction probability on the target class $y$, which can be formulated as:
\begin{equation}
    \frac{\partial \mathcal{L_\text{GCE}}(p,y)}{\partial \theta} = p^q_y\frac{\partial \mathcal{L_\text{CE}}(p,y)}{\partial \theta}.
\end{equation} 

The GCE loss encourages the model to focus on the easy samples with high probability values. 
As revealed in the work of Nam~\textit{et al.}~\shortcite{nam2020learning}, the bias attributes are easy to learn compared to the intrinsic attributes, so a model predicts bias-aligned samples with high probability values.
Due to this fact, in a biased dataset, the GCE loss encourages the model to focus mainly on the bias-aligned samples, leading the model to be biased. 

\noindent \textbf{Reweighting-based approaches. \enskip}
Recent state-of-the-art debiasing methods~\cite{nam2020learning, disentangled} reweight data samples by utilizing two different models: 1) a biased model $f_B$ and 2) a debiased model $f_D$.
The former one is trained to be overfitted to the bias attribute while the latter one is mainly trained with the bias-conflicting samples, those which are identified by utilizing $f_B$. 
To be more specific, since $f_B$ heavily relies on the bias attributes for making predictions, it fails to correctly classify the bias-conflicting samples, those without the bias attributes.
Due to this fact, the Cross Entropy (CE) loss values of bias-conflicting samples are relatively high compared to those of bias-aligned ones.
By utilizing such a characteristic, the loss of each data sample $x$ is reweighted for training $f_D$ with the reweighting value $w(x)$. 
Specifically, Nam~\textit{et al.}~\shortcite{nam2020learning} formulated $w(x)$ as
\begin{equation}
    w(x) = \frac{\mathcal{L_\text{CE}}(f_B(x), y)}{\mathcal{L_\text{CE}}(f_B(x), y) + \mathcal{L_\text{CE}}(f_D(x), y)},
\end{equation} 
where $f_B(x)$ and $f_D(x)$ indicate the prediction outputs of $f_B$ and $f_D$, respectively, and $y$ is the target label of the sample $x$.
Using the formula, the reweighting value $w(x)$ is designed to be imposed 1) high for the bias-conflicting samples and 2) low for the bias-aligned samples in order to improve the debiasing performance of $f_D$. 
In this regard, how well $f_B$ is overfitted to the bias attribute determines the $w(x)$ which crucially influences the debiasing performance of $f_D$. 

\begin{figure*}[t]
    \centering
    \includegraphics[width=0.75\textwidth]{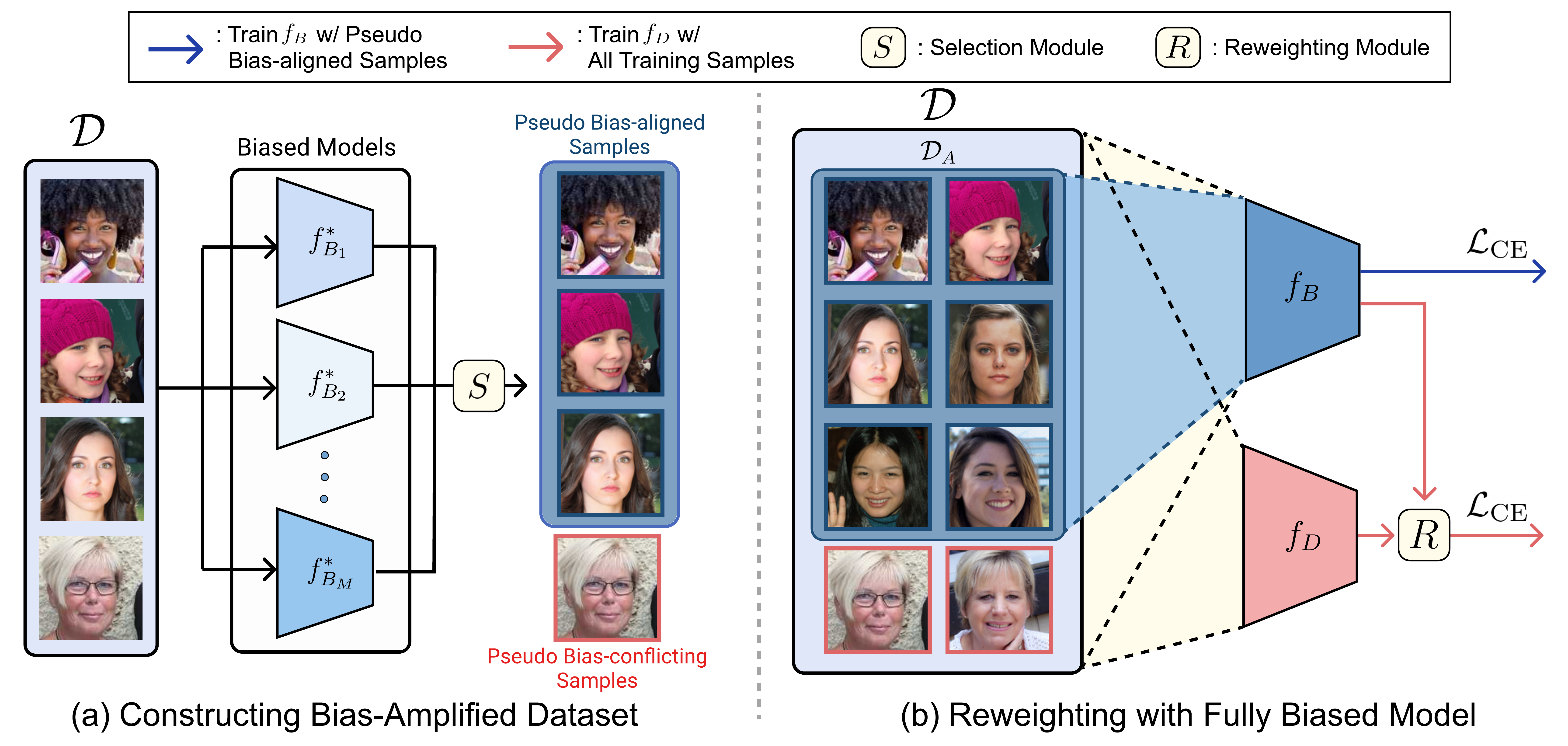}
    \vspace{-0.2cm}
    \caption{Illustration of BiasEnsemble. (a)
    By filtering out pseudo bias-conflicting samples detected via utilizing $M$ pretrained biased models~$(f_{B_1}^*,\dots,f_{B_M}^*)$, we obtain the bias-amplified dataset $\mathcal{D}_A$.
    (b) Then, we train $f_B$ with $\mathcal{D}_A$ while training $f_D$ with the original training set $\mathcal{D}$.
    $S$ and $R$ indicate the sample selection module and reweighting module, respectively.
    Although not used for training $f_B$, the pseudo bias-conflicting samples are still fed to $f_B$ for obtaining the reweighting value used for training $f_D$.
    }
    \vspace{-0.6cm}
    \label{fig:overview}
\end{figure*}

\vspace{-0.1cm}
\subsection{Revisiting $f_B$ in Debiasing Methods}
\label{sec:motivation-reweighting}
In this section, we show that the existing state-of-the-art reweighting methods fail to fully overfit $f_B$ to the bias, resulting in an unsatisfactory reweighting of the data samples during training $f_D$ overall. 
For the experiments, we use Colored MNIST~\cite{disentangled} and biased FFHQ (BFFHQ)~\cite{biaswap} to demonstrate that our analysis is applicable both on synthetic and real-world datasets.
Bias-conflicting samples consist 1\% of both training sets in this analysis.
Detailed descriptions of the datasets are included in Supplementary.
In Fig.~\ref{fig:reweighting-motivation}, we compare 1) LfF~\cite{nam2020learning} training $f_B$ with GCE loss (blue) and 2) LfF training $f_B$ with explicit bias labels using Cross Entropy (CE) loss (red).
For the evaluation, we use 1) a biased test set, a dataset having a similar data distribution as the biased training set, and 2) the unbiased test set, a dataset which has no correlation found in the biased training set.

\noindent \textbf{Imperfectly biased $f_B$. \enskip} 
A fully biased $f_B$ is likely to achieve 1) high accuracy on the biased test set and 2) low accuracy on the unbiased test set since it only uses the bias attribute as the visual cue for predictions. 
In other words, the gap between the biased test set accuracy and the unbiased test set accuracy increases as $f_B$ focuses on the bias attribute.
As shown in Fig.~\ref{fig:reweighting-motivation}, however, $f_B$ trained with GCE loss (blue) shows relatively higher unbiased test set accuracy compared to $f_B$ trained with the explicit bias labels (red).
Assuming that $f_B$ trained with the explicit bias labels is perfectly overfitted to the bias attribute, such results demonstrate that $f_B$ trained with GCE loss is less overfitted to the bias. 
In other words, even the small number of bias-conflicting samples work as noisy samples for learning bias.

\noindent \textbf{Debiasing $f_D$ via $f_B$. \enskip} 
The reweighting value $w(x)$ determines the degree of how much $f_D$ should focus on a given sample $x$ during the training phase.
It is crucial to satisfy two conditions simultaneously for training a debiased classifier $f_D$: imposing 1) high $w(x)$ on the bias-conflicting samples and 2) low $w(x)$ on the bias-aligned samples. 
In other words, the difference between $w(x)$ of bias-conflicting samples and that of bias-aligned samples, $w(x)_\text{diff}$, should be large in order to improve the debiasing performance. 
We observe that LfF trained with GCE loss, however, outputs relatively small $w(x)_\text{diff}$ compared to LfF trained with explicit bias labels (third and fourth column in Fig.~\ref{fig:reweighting-motivation}).
This is due to utilizing a less overfitted $f_B$ for computing $w(x)$.
Low $w(x)$ on bias-conflicting samples indicates that they are less emphasized in training $f_D$, which should be emphasized for learning debiased features (fourth column in Fig.~\ref{fig:reweighting-motivation}).
Therefore, $f_D$ of LfF trained with GCE loss shows lower test accuracy than the one trained with explicit bias labels (fifth column in Fig.~\ref{fig:reweighting-motivation}).
Based on the finding that how well $f_B$ is overfitted to the bias significantly influences the debiasing performance of $f_D$, we propose an approach which further amplifies the bias for training $f_B$. 

\vspace{-0.25cm}
\section{Debiasing with Bias-Amplified Dataset}

\subsection{Detecting Bias-conflicting Samples}
\label{subsec:bcd}
At a high level, since even a small number of bias-conflicting samples work as noisy samples for learning the bias, we discard them and build a bias-amplified dataset, mainly consisted of bias-aligned ones.
We pretrain an additional biased model $f_B^*$ with GCE loss for \emph{a small number of iterations} by utilizing the property that the bias attribute is easy to learn in the early training phase~\cite{nam2020learning}.
Note that $f^{*}_B$ is a pretrained biased model while $f_B$ is the biased model used for reweighting data samples during training $f_D$.
Since we pretrain $f_B^*$ only for a small number of iterations, our method requires a minimal amount of additional computational costs.

As $f_B^*$ is overfitted to the bias at a certain degree, it mainly 1) correctly classifies the bias-aligned samples and 2) misclassifies the bias-conflicting samples.
In other words, the model outputs 1) high confidence (\textit{i.e.,} the softmax probability) on the target class for the bias-aligned samples and 2) low confidence for the bias-conflicting samples. 
By utilizing the probability of the target class $p_y$, we build the bias-conflicting detector \emph{BCD} as follows:
\begin{equation}
    BCD(x; \tau, f)=
    \begin{cases}
    0, & \mbox{if } p_y(x;f) < \tau \\
    1, & \mbox{if } p_y(x;f) \ge \tau 
    \end{cases},
    \label{eq:align detector}
\end{equation}
where $\tau$ is the confidence threshold. 
The detector regards the samples with confidence higher than the threshold as bias-aligned samples and vice versa.

\vspace{-0.1cm}
\subsection{Improving Detection via Multiple BCDs}

While a single BCD may discard the bias-conflicting samples at a reasonable level, we empirically found that constructing $\mathcal{D}_A$ relying on only a single BCD shows large performance variations (Table~\ref{tab:ablation_voting} and Table~\ref{tab:ablation_be_amplified_dataset}). 
Although the bias attribute (\textit{e.g.,} gender bias) is easy to learn compared to the intrinsic attribute in the early training phase, it may form as a combination of multiple visual attributes (\textit{e.g.,} make-up, hairstyle, beards), especially in the real-world datasets. 
As shown in Fig.~\ref{fig:gradcam}, differently initialized biased models only utilize certain visual attributes for learning the bias attribute. 
Also, the visual attributes utilized for the biased predictions are different among models.
For example, one BCD captures the gender bias of a female image by mainly using the long hair as the visual cue while the other may recognize the bias mainly due to the lip makeups.
Thus, each BCD may make different predictions on a same sample, leading to performance variation overall.
One of the straight-forward solutions is considering both visual attributes to predict gender bias (\textit{i.e.,} predicting an image as female if it includes both long hair and lip makeups).
As demonstrated in the previous studies~\cite{ensemble-help-generalization, ensemble-loss-landscape, ensemble-search-uncertainty}, utilizing differently initialized models enables to induce diversity among models.
Thus, we utilize multiple BCDs to better capture the bias via considering diverse visual attributes consisting the bias attribute.
Since we utilize multiple BCDs, we term our method as `BiasEnsemble (BE)'.

To this end, we select data samples based on the predictions of multiple BCDs.
To be more specific, we leverage multiple pretrained biased models~($f_{B_1}^*,f_{B_2}^*,...,f_{B_M}^*$).
We utilize the property that bias attribute is learned in the \emph{early training phase}~\cite{nam2020learning}, so we only need a negligible training time for each $f_{B}^{*}$.
Quantitative measurement on the marginal computational costs of our method is reported in our Supplementary.
Then, $M$ number of BCDs are built using each pretrained biased model $f_B^*$.
Finally, we discard the sample that the majority of the detectors consider as the bias-conflicting sample. 
For example, setting $M$=$5$, a given sample is regarded as the bias-conflicting sample if more or equal to three BCDs considered it as the bias-conflicting one (\textit{i.e.,} pseudo bias-conflicting sample) and vice versa. 
Note that all biased models have the same architecture, so we iteratively re-initialize biased models in order to save the memory space. 
As aforementioned, even such iterative re-initialization accompanies a marginal training time since we train each biased model for a small number of steps.

In summary, pseudo bias-aligned sample (PBA) can be formulated as 
\vspace{-0.1cm}
\begin{equation}
    PBA = 
    \begin{cases} 
    0, & \mbox{if } \sum_{i=1}^M{BCD(x; \tau, f_{B_i}^*)} < \lceil{M\over2}\rceil\\
    1, & \mbox{if } \sum_{i=1}^M{BCD(x; \tau, f_{B_i}^*)} \ge \lceil{M\over2}\rceil 
    \end{cases},
    \label{eq:align selector}
\end{equation}
where $M$ is the number of BCDs used.
Finally, the data samples labeled as pseudo bias-aligned ones consist $\mathcal{D}_A$, used for training $f_B$. 

In the Supplementary, given that bias labels are not provided, we show that BiasEnsemble is superior to simply ensembling multiple biased models ($f_{B_1}$, $f_{B_2}$, ..., $f_{B_M}$) in the main stage of debiasing.
Without the bias-conflicting samples discarded, the ensembled predictions of the multiple biased models fail to emphasize the bias-conflicting ones for $f_D$.
That is, each biased model ($f_{B_i}$) learns the intrinsic attribute from the bias-conflicting samples as training proceeds. 
While ensembling has been widely adopted in other fields to bring further performance gain, this experiment shows that ensembling without careful consideration does not guarantee performance gain in debiasing. 
Although ensembling itself may be regarded as a simple and naive approach, we believe that finding how to adjust ensembling to debiasing is important and needs careful consideration.

\vspace{-0.1cm}
\subsection{Training Debiased Model Using $\mathcal{D}_A$}
\label{sec:train_using_be}
\vspace{-0.1cm}
After obtaining a bias-amplified dataset $\mathcal{D}_A$ which contains a significantly smaller number of bias-conflicting samples compared to the original training dataset $\mathcal{D}$, we train $f_B$ using $\mathcal{D}_A$.
When applying BiasEnsemble to existing reweighting-based approaches, LfF~\cite{nam2020learning} and DisEnt~\cite{disentangled}, we do not modify the training procedure of $f_D$. Thus, BiasEnsemble can be easily applied to existing methods that leverage $f_B$ for \emph{reweighting} data samples.
Note that $f_B$ is utilized for reweighting all the training data samples during training $f_D$, although the pseudo bias-conflicting ones (\textit{i.e.,} the samples not included in $\mathcal{D}_A$) are not used for \emph{training} $f_B$. 
Both $f_B$ and $f_D$ are trained with the CE loss. 

\vspace{-0.2cm}
\section{Experiment}
\begin{table*}[t!]
\small
\begin{center}
\resizebox{\textwidth}{!}{%
\large
\setlength{\tabcolsep}{0.2em}
\def\arraystretch{1.5}%
\begin{tabular}{cc|cccc|cccc|cc|cc}
\toprule
\multicolumn{2}{c}
{\multirow{2}{*}{Method}}
& \multicolumn{4}{c}{Colored MNIST}
& \multicolumn{4}{c}{BFFHQ}
& \multicolumn{2}{c}{Dogs \& Cats}
& \multicolumn{2}{c}{BAR}
\\ \cmidrule(lr){3-6}\cmidrule(lr){7-10}\cmidrule(lr){11-12}\cmidrule(lr){13-14} 
\multicolumn{2}{c}{}
& 0.5\% & 1.0\% & 2.0\% & 5.0\% & 0.5\% & 1.0\% & 2.0\% & 5.0\% & 1.0\% & 5.0\% & 1.0\% & 5.0\% \\
\midrule
Vanilla~\cite{He2015resnet} & \textcolor{black}{\boldxmark} \textcolor{black}{\boldxmark} & 34.75  & 51.14  & 65.72  & 82.82  & 55.64  & 60.96  & 69.00  & 82.88  & 48.06  & 69.88  & 70.55  & 82.53 \\
HEX~\cite{wang2018hex} & \textcolor{black}{\boldxmark} \textcolor{black}{\boldcheckmark} & 42.25  & 47.02  & 72.82  & 85.50  & 56.96  & 62.32  & 70.72  & 83.40  & 46.76  & 72.60  & 70.48  & 81.20 \\
LNL~\cite{LNL} & \textcolor{black}{\boldcheckmark} \textcolor{black}{\boldcheckmark} & 36.29  & 49.48  & 63.30  & 81.30  & 56.88  & 62.64  & 69.80  & 83.08  & 50.90  & 73.96  & - & - \\
EnD~\cite{EnD} & \textcolor{black}{\boldcheckmark} \textcolor{black}{\boldcheckmark} & 35.33  & 48.97  & 67.01  & 82.09  & 55.96  & 60.88  & 69.72  & 82.88  & 48.56  & 68.24  & -  & - \\
ReBias~\cite{bahng2019rebias} & \textcolor{black}{\boldxmark} \textcolor{black}{\boldcheckmark} & 60.86  & \textbf{82.78}  & \textbf{92.00}  & \textbf{96.45}  & 55.76  & 60.68  & 69.60  & 82.64  & 48.70  & 65.74  & 73.04  & 83.90 \\
LfF~\cite{nam2020learning} & \textcolor{black}{\boldxmark} \textcolor{black}{\boldxmark} & 63.55  & 76.81  & 84.18  & 89.65  & 65.19  & 69.24  & 73.08  & 79.80  & 71.72  & 84.32  & 70.16  & 82.95 \\
DisEnt~\cite{disentangled} & \textcolor{black}{\boldxmark} \textcolor{black}{\boldxmark} & 68.49  & 79.99  & 84.09  & 89.91  & 62.08  & 66.00  & 69.92  & 80.68  & 65.74  & 81.58  & 70.33  & 83.13 \\
\midrule
\multirow{2}{*}{LfF + BE} & \multirow{2}{*}{\textcolor{black}{\boldxmark} \textcolor{black}{\boldxmark}} & 69.70 & 81.17 & 85.20 & 90.04 & 67.36 & \textbf{75.08} & \textbf{80.32} & \textbf{85.48} & \textbf{81.52} & \textbf{88.60} & \textbf{73.36} & 83.87 \\
&  & \cellcolor{Gray} (+ 6.15) & \cellcolor{Gray} (+ 4.36) & \cellcolor{Gray} (+ 1.02) & \cellcolor{Gray} (+ 0.39) & \cellcolor{Gray} (+ 2.17) & \cellcolor{Gray} (+ 5.84) & \cellcolor{Gray} (+ 7.24) & \cellcolor{Gray} (+ 5.68) & \cellcolor{Gray} (+ 9.80) & \cellcolor{Gray} (+ 4.28) & \cellcolor{Gray} (+ 3.20) & \cellcolor{Gray} (+ 0.92) \\
\multirow{2}{*}{DisEnt + BE} & \multirow{2}{*}{\textcolor{black}{\boldxmark} \textcolor{black}{\boldxmark}} & \textbf{71.34} & 82.11 & 84.66 & 90.15 & \textbf{67.56} & 73.48 & 79.48 & 84.84 & 80.74 & 86.84 & 73.29 & \textbf{84.96} \\
&  & \cellcolor{Gray} \cellcolor{Gray} (+ 2.85) & \cellcolor{Gray} (+ 2.12) & \cellcolor{Gray} (+ 0.57) & \cellcolor{Gray} (+ 0.24) & \cellcolor{Gray} (+ 5.48) & \cellcolor{Gray} (+ 7.48) & \cellcolor{Gray} (+ 9.56) & \cellcolor{Gray} (+ 4.16) & \cellcolor{Gray} (+ 15.00) & \cellcolor{Gray} (+ 5.26) & \cellcolor{Gray} (+ 2.96) & \cellcolor{Gray} (+ 1.83) \\

\bottomrule
\end{tabular}%
}
\vspace{-0.25cm}
\caption{Image classification accuracy on unbiased test sets with varying ratios of bias-conflicting samples.
The \textit{cross} and \textit{check} represent whether each model 1) uses bias labels during training and 2) requires predefined bias type.
For LfF and DisEnt, the performance gains are shaded in grey.
Best performing results are marked in bold.}
\vspace{-0.5cm}
\label{tab:best_test}
\end{center}
\end{table*}

\label{sec:experiment-total}

\subsection{Experimental Settings}
\label{sec:experiment}
\vspace{-0.05cm}
\noindent{\textbf{Dataset. \enskip}}
Following the previous studies, we conduct experiments under four datasets: Colored MNIST~\cite{disentangled}, biased FFHQ (BFFHQ)~\cite{biaswap}, Dogs \& Cats~\cite{LNL}, and biased action recognition (BAR)~\cite{nam2020learning}.
Each dataset has an intrinsic attribute and a bias attribute: Colored MNIST - \{digit, color\}, BFFHQ - \{age, gender\}, Dogs \& Cats - \{animal, color\}, and BAR - \{action, background\}.
The former and the latter visual attribute in the bracket correspond to the intrinsic and bias attribute, respectively.
We conduct experiments under various ratios of bias-conflicting samples (\textit{i.e.,} the number of bias-conflicting samples out of the total number of training samples) in each dataset to evaluate the debiasing algorithms under different levels of bias severity, following the previous studies~\cite{nam2020learning, disentangled}.
For evaluating the debiasing performance, we use unbiased test sets which include images without the correlation found in the training set.
We use datasets with 1\% ratio of bias-conflicting samples for in-depth analyses.

\begin{figure*}[t]
    \centering
    \vspace{-0.1cm}
    \includegraphics[width=0.85\textwidth]{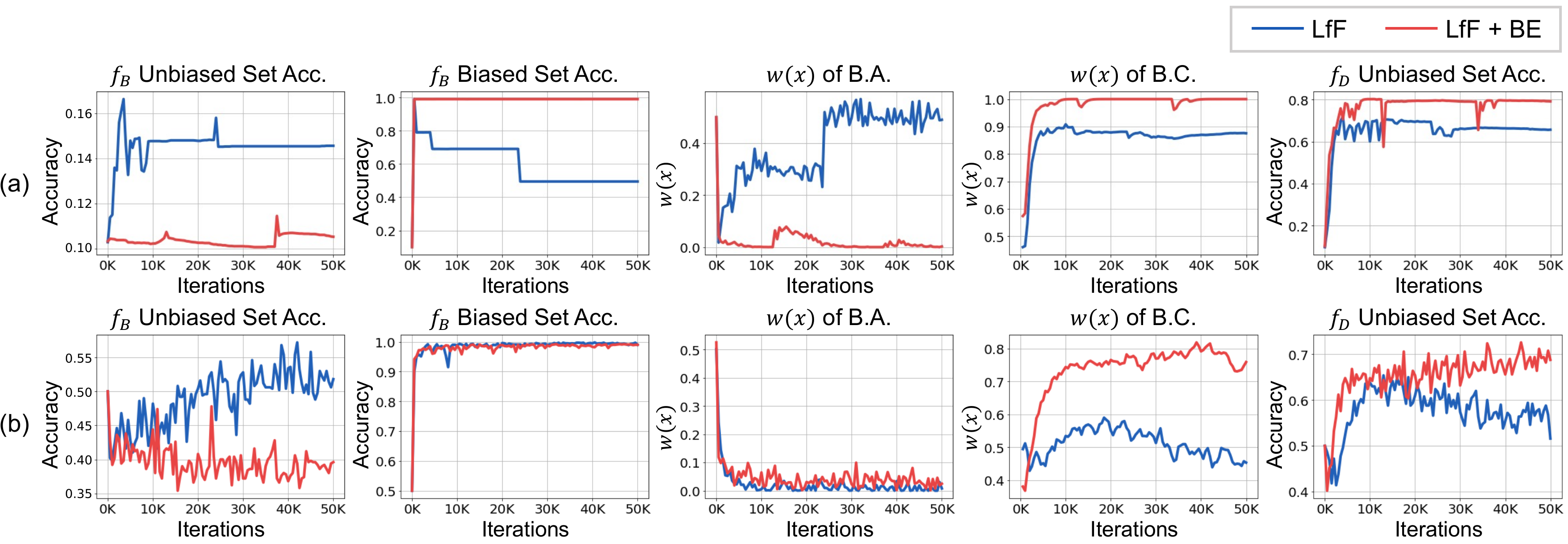}
    \vspace{-0.3cm}
    \caption{
    Comparison of LfF trained 1) without BiasEnsemble (blue) and 2) with BiasEnsemble (red) on (a) Colored MNIST and (b) BFFHQ. 
    Each column corresponds to the ones in Fig.~\ref{fig:reweighting-motivation}.}
    \vspace{-0.65cm}
    \label{fig:reweighting-ours}
\end{figure*}

\noindent{\textbf{Implementation details. \enskip}}
Following Nam~\textit{et al.}~\shortcite{nam2020learning} and Lee~\textit{et al.}~\shortcite{disentangled}, we use a multi-layer perceptron~(MLP) which consists of three hidden layers for Colored MNIST.
For the other datasets except for BAR, we train ResNet18~\cite{He2015resnet} with the random initialization.
Since BAR has an extremely small number of images compared to other datasets, we utilize a pretrained ResNet18.
We set $M$=$5$, meaning that we pretrain five biased models~(\textit{i.e.,} $f^{*}_{B_1}, f^{*}_{B_1}, \dots, f^{*}_{B_5}$).
While all experiments are trained for 50K iterations, each $f^{*}_{B}$ is pretrained for 1K iterations on all datasets, requiring negligible amount of additional computational costs.
We set the confidence threshold $\tau$ for the BCD as 0.99.
Note that all the hyper-parameters are constant across all datasets and bias ratios.
We report the mean of the best unbiased test set accuracy over five independent trials.
We include the remaining details of datasets and implementation in the Supplementary. 

\vspace{-0.15cm}
\subsection{Comparisons on Unbiased Test Sets}
\label{sec:quantitative}
\vspace{-0.05cm}
Table~\ref{tab:best_test} compares the image classification accuracies of the debiasing approaches on the unbiased test sets. 
As aforementioned, we applied BiasEnsemble on the state-of-the-art reweighting-based approaches, LfF~\cite{nam2020learning} and DisEnt~\cite{disentangled}. 
We found that using BiasEnsemble for the two methods significantly improves the debiasing performances in four datasets regardless of the bias severities.
The performance gains are colored grey in Table~\ref{tab:best_test}. 
We also observe that applying BiasEnsemble brings larger performance gain when evaluated with real-world datasets compared to the synthetic dataset.
For example, using BiasEnsemble on DisEnt shows 7.48\% and 9.56\% performance gain on BFFHQ with 1\% and 2\% ratio of bias-conflicting samples, respectively.
Note that we could not evaluate the debiasing methods requiring bias labels (LNL and EnD) on BAR dataset since the dataset does not include explicit bias labels.

Utilizing our approach on DisEnt outperforms ReBias~\cite{bahng2019rebias} on BAR.
BAR dataset is biased towards the background which mainly contains the color and texture bias.
Since ReBias uses BagNet~\cite{brendel2018bagnets} which is a color- and texture-oriented model to identify bias (\textit{i.e.,} leveraging a prior knowledge on the bias type), it showed the state-of-the-art performance before using BiasEnsemble on existing reweighting-based approaches. 
However, even without such prior knowledge on the bias type, applying BiasEnsemble on DisEnt outperforms ReBias regardless of bias severities. 

For the Colored MNIST, ReBias utilizes four layers of convolutional neural network while the other debiasing methods use three layers of multi-layer perceptron.
We inevitably use the convolutional neural network for ReBias since it leverages a small receptive field of convolutional layers to capture the color bias. 
When comparing with the baselines using the same architecture, leveraging BiasEnsemble on LfF and DisEnt achieve the state-of-the-art debiasing performance on Colored MNIST.

\vspace{-0.2cm}
\subsection{Analysis}
\label{sec:analysis}

\noindent \textbf{Amplified bias of $f_B$. \enskip}
Similar to Fig.~\ref{fig:reweighting-motivation} which describes the motivation of our work, we compare LfF trained 1) without BiasEnsemble and 2) with BiasEnsemble in Fig.~\ref{fig:reweighting-ours}.
While achieving comparable or higher biased test set accuracy, $f_B$ with BiasEnsemble shows lower unbiased test set accuracy compared to $f_B$ without applying BiasEnsemble. 
This leads to increase the $w(x)_\text{diff}$, the difference between $w(x)$ of bias-conflicting samples and that of bias-aligned ones.
Then, bias-conflicting samples are further emphasized for training $f_D$, improving the debiasing performance overall.
Such improvement is valid in both synthetic (\textit{i.e.,} Colored MNIST) and the real-world dataset (\textit{i.e.,} BFFHQ). 
This visualization demonstrates that our proposed method indeed improves debiasing performance of $f_D$ by further amplifying bias of $f_B$.

\noindent \textbf{How to construct bias-amplified $\mathcal{D}_A$ for debiasing. \enskip}
We found two important factors when constructing $\mathcal{D}_A$: 1) discarding sufficient number of bias-conflicting samples and 2) maintaining a reasonable number of bias-aligned ones.
To understand how the data samples composing $\mathcal{D}_A$ affects the debiasing performance, Table~\ref{tab:ablation_align_conflict} compares the debiasing performances of LfF by adjusting the number of bias-aligned and bias-conflicting samples in $\mathcal{D}_A$.
In Table~\ref{tab:ablation_align_conflict}, \# of B.A. and \# of B.C. indicate the remaining number of bias-aligned samples and that of bias-conflicting samples in $\mathcal{D}_A$, respectively, computed in ratio compared to the original training set.  
For example, in the case of adjusted ratios of (20\%, 20\%), 50000 bias-aligned samples and 100 bias-conflicting samples in $\mathcal{D}$ are adjusted to 10000 and 20 in $\mathcal{D}_A$, respectively. 
We trained $f_B$ by using the adjusted dataset while using the original training set for training $f_D$.

\begin{table}[t]
\vspace{-0.3cm}
\centering
\resizebox{0.8\linewidth}{!}{
\setlength{\tabcolsep}{0.5em}
\def\arraystretch{1.0}%
\begin{tabular}{c | c c c c c}
\toprule 

\# of B.A.          & 100\%   &  100\%    & 100\%    & 60\%   & 20\%   \\
\# of B.C.          & 100\%   &  60\%     & 20\%     & 20\%   & 20\%   \\
\midrule
Colored MNIST       & 58.48   &  73.87     & 81.58    & 75.75   & 63.21   \\
BFFHQ               & 62.10   &  73.96     & 79.36    & 75.88   & 69.12   \\
Dogs\&Cats          & 53.10   &  71.00     & 79.22    & 63.86   & 60.78   \\

\bottomrule
\end{tabular}}
\vspace{-0.2cm}
\caption{
Unbiased test set accuracies with adjusted number of bias-aligned samples~(\# of B.A.) and that of bias-conflicting ones~(\# of B.C.) in the bias-amplified dataset $\mathcal{D}_A$ utilized for training $f_B$ of LfF with CE loss.
First two rows represent the ratio of samples in $\mathcal{D}_A$ compared to $\mathcal{D}$. 
}
\vspace{-2.2mm}
\label{tab:ablation_align_conflict}
\end{table}
\begin{table}[t]
\centering
\resizebox{1.0\linewidth}{!}{
\begin{tabular}{c | c | c c c c}
\toprule 

Method & $M$ & Colored MNIST & BFFHQ & Dogs \& Cats & BAR \\
\midrule
LfF & - & 76.81\stdv{4.56} & 69.24\stdv{2.07} & 71.72\stdv{4.56} & 70.16\stdv{0.77} \\
\midrule
\multirow{2}{*}{LfF + BE} & 1 & 79.51\stdv{1.56} & 71.52\stdv{2.68} & 76.98\stdv{6.63} & 71.63\stdv{1.59} \\
                        & 5 & \textbf{81.17}\stdv{0.68} & \textbf{75.08}\stdv{2.29} & \textbf{81.52}\stdv{1.13} & \textbf{73.36}\stdv{0.97} \\

\bottomrule
\end{tabular}}
\vspace{-0.2cm}
\caption{
    Unbiased test set accuracies on 1) LfF, 2) applying our method on LfF with a single BCD and 3) multiple BCDs. 
    $M$ indicates the number of BCDs when using our method.
}
\vspace{-0.6cm}
\label{tab:ablation_voting}
\end{table}

When fixing the number of bias-aligned samples constant ($100\%$), the debiasing performance improves as the number of bias-conflicting samples decreases (from $100\%$ to $20\%$).
The main reason is that the bias-conflicting samples, preventing $f_B$ from learning the bias attribute, are discarded.
This demonstrates that discarding sufficient number of bias-conflicting samples is important for improving the debiasing performance which is straight-forward.
On the other hand, we also observe that debiasing performance deteriorates when the number of bias-aligned samples decreases (from $100\%$ to $20\%$) with the constant number of bias-conflicting samples ($20\%$).
This indicates that $f_B$ also requires a sufficient number of bias-aligned samples to learn the bias attributes.
We want to emphasize that simply discarding numerous number of training samples for the purpose of eliminating entire bias-conflicting samples may fail to bring large performance gain since it also filters out bias-aligned ones, those important for learning a bias.
This analysis demonstrates the importance of considering both factors when constructing $\mathcal{D}_A$.
In the Supplementary, along with the standard deviation, we gradually change the adjusted number of samples (\textit{e.g.,} 80\%, 40\%) to show the detailed tendency of change in debiasing performance with respect to the adjusted number of samples.

\noindent \textbf{Superiority of multiple BCDs over single BCD. \enskip}
We compare the debiasing performance of using BiasEnsemble with a single BCD ($M$=$1$) and multiple BCDs ($M$=$5$) on LfF in Table~\ref{tab:ablation_voting}.
While using a single BCD brings performance gain compared to LfF, the standard deviation of the performance is larger when compared to using multiple BCDs.
For example, using a single BCD to train LfF shows the standard deviation of 6.63\% on Dogs \& Cats dataset. 
This is due to the fact that we rely on a single BCD for constructing $\mathcal{D}_A$. 
When the single BCD fails to be overfitted to the bias, it fails to filter out the bias-conflicting samples for building $\mathcal{D}_A$.
However, such an issue is mitigated when using multiple BCDs since they better capture the bias attribute by considering multiple visual attributes of the bias, compared to using a single BCD. 
We provide the further analysis on the performance variations of single BCD and multiple BCDs in Supplementary.

Such result is mainly due to the number of bias-aligned samples and bias-conflicting ones included in $\mathcal{D}_A$, as demonstrated in Table~\ref{tab:ablation_align_conflict}.
Table~\ref{tab:ablation_be_amplified_dataset} shows the remaining number of bias-aligned samples and bias-conflicting samples in $\mathcal{D}_A$ after applying BiasEnsemble, each computed in ratio compared to the original training dataset $\mathcal{D}$.
We observe that utilizing multiple BCDs 1) maintains a significant number of bias-aligned samples and 2) further reduces the number of bias-conflicting samples compared to using a single BCD. 
Additionally, the standard deviations of the remaining number of samples are considerably larger when using the single BCD, demonstrating that a single BCD fails to fully capture the bias attribute at a stable level.

\begin{table}[t!]
\centering
\vspace{-0.4cm}
\resizebox{0.95\linewidth}{!}{
\setlength{\tabcolsep}{0.5em}
\begin{tabular}{ c | c c | c c}
\toprule 


\multicolumn{1}{c}{}& \multicolumn{2}{c}{\# of B.A.(\%)$\uparrow$}
& \multicolumn{2}{c}{\# of B.C.(\%)$\downarrow$}
\\ \cmidrule(lr){2-3}\cmidrule(lr){4-5}
\multicolumn{1}{c}{Dataset} & M=1 & M=5 & M=1 & M=5 \\
\midrule
Colored MNIST & 84.10\stdv{11.01} &  \textbf{99.96}\stdv{0.03} &  4.64\stdv{0.79} & \textbf{1.42}\stdv{0.78} \\
BFFHQ & 84.51\stdv{4.34} &  \textbf{92.22}\stdv{0.26} &  24.47\stdv{4.63} & \textbf{24.37}\stdv{3.07}\\
Dogs\&Cats & 85.89\stdv{2.61} &  \textbf{88.60}\stdv{1.02} &  12.00\stdv{6.25} & \textbf{9.50}\stdv{3.75}\\
BAR & 97.24\stdv{0.27} &  \textbf{98.39}\stdv{0.19} &  60.00\stdv{13.24} & \textbf{51.42}\stdv{5.34}\\

\bottomrule
\end{tabular}}
\vspace{-0.2cm}
\caption{
The remaining number of bias-aligned samples~(\# of B.A.) and bias-conflicting ones~(\# of B.C.) in the bias-amplified dataset $\mathcal{D}_A$ after applying our method.
The remaining numbers are shown in ratios of samples compared to the original training dataset $\mathcal{D}$.}
\vspace{-0.4cm}
\label{tab:ablation_be_amplified_dataset}
\end{table}



\begin{table}[t]
\vspace{0.15cm}
\begin{center}
\label{table:headings}
\resizebox{1.0\linewidth}{!}{
\setlength{\tabcolsep}{0.7em}
\begin{tabular}{c|c|ccccc}
\toprule 


    

Method & LfF & \multicolumn{5}{c}{LfF + BE} \\
\midrule
Threshold & - & 0.9	& 0.95  & 0.98 & 0.99 & 0.999 \\
\midrule
Colored MNIST & 76.81 & 80.65 & 80.49 & 80.41  & \textbf{81.17} &    81.06 \\

BFFHQ         & 69.24 & 73.16 & 74.48 & 74.28  & \textbf{75.08} &	72.32 \\

Dogs \& Cats  & 71.72 & 80.60 & 77.66 & 80.36  & \textbf{81.52} &	77.90 \\
    
BAR           & 70.16 & 71.53 & 72.23 & 73.00  & \textbf{73.36} &	71.28 \\

\bottomrule
\end{tabular}}
\vspace{-0.3cm}
\caption{
    Unbiased test set accuracies evaluated with a wide range of confidence threshold $\tau$ of our method.
    }
\label{tab:ablation_threshold}
\vspace{-0.8cm}
\end{center}
\end{table}

\noindent \textbf{Robustness of BiasEnsemble across $\tau$.}
Table~\ref{tab:ablation_threshold} shows that BiasEnsemble is robust to the confidence threshold $\tau$. We report the image classification accuracy on the unbiased test set using a wide range of $\tau$ (\textit{i.e.,} from 0.9 to 0.999). We observe that the test set accuracies constantly improve by applying BiasEnsemble to LfF, regardless of the value of $\tau$.

\vspace{-0.2cm}
\section{Conclusion}
In this work, we propose a biased sample selection method, BiasEnsemble, in order to train $f_B$ to maximally exploit the bias attribute.
Our main finding is that how well $f_B$ is overfitted to the bias influences the debiasing performance of $f_D$ which was overlooked in the previous debiasing studies.
While training $f_B$ to overfit to the bias, the bias-conflicting samples interfere with learning bias for $f_B$, so we filter them out to construct a refined bias-amplified dataset $\mathcal{D}_A$.
To do so, we utilize differently randomly initialized biased models to consider diverse visual attributes to better capture the bias attribute and discard the bias-conflicting samples for constructing $\mathcal{D}_A$.
Such a simple approach improves the recent state-of-the-art reweighting-based debiasing approaches. 
We believe that we shed light on an important debiasing component $f_B$ which has been relatively overlooked compared to $f_D$, and provide insightful findings for future researchers in debiasing.

\section{Acknowledgments}

This work was supported by the Institute of Information \& communications Technology Planning \& Evaluation (IITP) grant funded
by the Korean government (MSIT) (No. 2019-0-00075 and No.2022-0-009840101003, Artificial Intelligence Graduate School Program (KAIST)), Electronics and Telecommunications Research Institute(ETRI) grant funded by the Korean government [22ZS1200, Fundamental Technology Research for Human-Centric Autonomous Intelligent Systems], and Kakao Enterprise.
\bibliography{aaai23}

\begin{thebibliography}{21}
\providecommand{\natexlab}[1]{#1}

\bibitem[{Bahng et~al.(2020)Bahng, Chun, Yun, Choo, and Oh}]{bahng2019rebias}
Bahng, H.; Chun, S.; Yun, S.; Choo, J.; and Oh, S.~J. 2020.
\newblock Learning De-biased Representations with Biased Representations.
\newblock In \emph{International Conference on Machine Learning (ICML)}.

\bibitem[{Bian and Chen(2021)}]{ensemble-help-generalization}
Bian, Y.; and Chen, H. 2021.
\newblock When Does Diversity Help Generalization in Classification Ensembles?
\newblock \emph{IEEE Transactions on Cybernetics}, 1--17.

\bibitem[{Brendel and Bethge(2019)}]{brendel2018bagnets}
Brendel, W.; and Bethge, M. 2019.
\newblock Approximating CNNs with Bag-of-local-Features models works
  surprisingly well on ImageNet.
\newblock \emph{International Conference on Learning Representations}.

\bibitem[{Darlow, Jastrzebski, and Storkey(2020)}]{darlow2020latent}
Darlow, L.; Jastrzebski, S.; and Storkey, A. 2020.
\newblock Latent Adversarial Debiasing: Mitigating Collider Bias in Deep Neural
  Networks.
\newblock \emph{arXiv preprint arXiv:2011.11486}.

\bibitem[{Fort, Hu, and Lakshminarayanan(2019)}]{ensemble-loss-landscape}
Fort, S.; Hu, H.; and Lakshminarayanan, B. 2019.
\newblock Deep Ensembles: A Loss Landscape Perspective.

\bibitem[{Geirhos et~al.(2019)Geirhos, Rubisch, Michaelis, Bethge, Wichmann,
  and Brendel}]{geirhos2018imagenettrained}
Geirhos, R.; Rubisch, P.; Michaelis, C.; Bethge, M.; Wichmann, F.~A.; and
  Brendel, W. 2019.
\newblock ImageNet-trained {CNN}s are biased towards texture; increasing shape
  bias improves accuracy and robustness.
\newblock In \emph{International Conference on Learning Representations}.

\bibitem[{He et~al.(2015)He, Zhang, Ren, and Sun}]{He2015resnet}
He, K.; Zhang, X.; Ren, S.; and Sun, J. 2015.
\newblock Deep Residual Learning for Image Recognition.
\newblock \emph{arXiv preprint arXiv:1512.03385}.

\bibitem[{Huang et~al.(2020)Huang, Wang, Xing, and Huang}]{huangRSC2020}
Huang, Z.; Wang, H.; Xing, E.~P.; and Huang, D. 2020.
\newblock Self-Challenging Improves Cross-Domain Generalization.
\newblock In \emph{ECCV}.

\bibitem[{Karras, Laine, and Aila(2019)}]{stylegan}
Karras, T.; Laine, S.; and Aila, T. 2019.
\newblock A Style-Based Generator Architecture for Generative Adversarial
  Networks.
\newblock In \emph{Proceedings of the IEEE/CVF Conference on Computer Vision
  and Pattern Recognition (CVPR)}.

\bibitem[{Kim et~al.(2019)Kim, Kim, Kim, Kim, and Kim}]{LNL}
Kim, B.; Kim, H.; Kim, K.; Kim, S.; and Kim, J. 2019.
\newblock Learning Not to Learn: Training Deep Neural Networks With Biased
  Data.
\newblock In \emph{The IEEE Conference on Computer Vision and Pattern
  Recognition (CVPR)}.

\bibitem[{Kim, Lee, and Choo(2021)}]{biaswap}
Kim, E.; Lee, J.; and Choo, J. 2021.
\newblock BiaSwap: Removing Dataset Bias With Bias-Tailored Swapping
  Augmentation.
\newblock In \emph{Proceedings of the IEEE/CVF International Conference on
  Computer Vision (ICCV)}, 14992--15001.

\bibitem[{LeCun and Cortes(2010)}]{mnist}
LeCun, Y.; and Cortes, C. 2010.
\newblock {MNIST} handwritten digit database.

\bibitem[{Lee et~al.(2021)Lee, Kim, Lee, Lee, and Choo}]{disentangled}
Lee, J.; Kim, E.; Lee, J.; Lee, J.; and Choo, J. 2021.
\newblock Learning Debiased Representation via Disentangled Feature
  Augmentation.
\newblock In \emph{Advances in Neural Information Processing Systems}.

\bibitem[{Nam et~al.(2020)Nam, Cha, Ahn, Lee, and Shin}]{nam2020learning}
Nam, J.; Cha, H.; Ahn, S.; Lee, J.; and Shin, J. 2020.
\newblock Learning from Failure: Training Debiased Classifier from Biased
  Classifier.
\newblock In \emph{Advances in Neural Information Processing Systems}.

\bibitem[{Or-El et~al.(2020)Or-El, Sengupta, Fried, Shechtman, and
  Kemelmacher-Shlizerman}]{or2020lifespan}
Or-El, R.; Sengupta, S.; Fried, O.; Shechtman, E.; and Kemelmacher-Shlizerman,
  I. 2020.
\newblock Lifespan age transformation synthesis.
\newblock In \emph{European Conference on Computer Vision}, 739--755. Springer.

\bibitem[{Sagawa et~al.(2020)Sagawa, Koh, Hashimoto, and
  Liang}]{sagawa2019distributionally}
Sagawa, S.; Koh, P.~W.; Hashimoto, T.~B.; and Liang, P. 2020.
\newblock Distributionally robust neural networks for group shifts: On the
  importance of regularization for worst-case generalization.
\newblock In \emph{International Conference on Learning Representations
  (ICLR)}.

\bibitem[{Tartaglione, Barbano, and Grangetto(2021)}]{EnD}
Tartaglione, E.; Barbano, C.~A.; and Grangetto, M. 2021.
\newblock EnD: Entangling and Disentangling Deep Representations for Bias
  Correction.
\newblock In \emph{Proceedings of the IEEE/CVF Conference on Computer Vision
  and Pattern Recognition (CVPR)}, 13508--13517.

\bibitem[{Torralba and Efros(2011)}]{unbiaslook2011torralba}
Torralba, A.; and Efros, A.~A. 2011.
\newblock Unbiased look at dataset bias.
\newblock 1521--1528. IEEE Computer Society.

\bibitem[{Wang et~al.(2019)Wang, He, Lipton, and Xing}]{wang2018hex}
Wang, H.; He, Z.; Lipton, Z.~L.; and Xing, E.~P. 2019.
\newblock Learning Robust Representations by Projecting Superficial Statistics
  Out.
\newblock In \emph{International Conference on Learning Representations}.

\bibitem[{Zaidi et~al.(2021)Zaidi, Zela, Elsken, Holmes, Hutter, and
  Teh}]{ensemble-search-uncertainty}
Zaidi, S.; Zela, A.; Elsken, T.; Holmes, C.~C.; Hutter, F.; and Teh, Y. 2021.
\newblock Neural Ensemble Search for Uncertainty Estimation and Dataset Shift.
\newblock In \emph{Advances in Neural Information Processing Systems},
  volume~34.

\bibitem[{Zhang and Sabuncu(2018)}]{zhang2018generalized}
Zhang, Z.; and Sabuncu, M.~R. 2018.
\newblock Generalized cross entropy loss for training deep neural networks with
  noisy labels.
\newblock \emph{arXiv preprint arXiv:1805.07836}.

\end{thebibliography}

\clearpage

\appendix

This supplementary presents additional experiments and further explanations of our approach that we could not include in the main paper due to the page limit. 




\section{Comparisons with Ensembles}

\begin{table*}[t!]
\centering
\resizebox{0.75\linewidth}{!}{
\setlength{\tabcolsep}{0.7em}
\begin{tabular}{ c | c | c c c c }
\toprule 
Dataset       & LfF               & LfF ($f_B$ Ensemble) & LfF ($f_B$ Soft BE) & LfF ($f_D$ Ensemble) & LfF + BE              \\
\midrule
Colored MNIST & 76.81 \stdv{4.56} & 66.73 \stdv{2.87} & 80.63 \stdv{2.29} & 78.71 \stdv{2.70} & \textbf{81.17} \stdv{0.68} \\
BFFHQ         & 69.24 \stdv{2.07} & 69.14 \stdv{1.88} & 73.36 \stdv{1.10} & 70.49 \stdv{1.14} & \textbf{75.08} \stdv{2.29} \\
Dogs \& Cats  & 71.72 \stdv{4.56} & 71.90 \stdv{1.70} & 81.20 \stdv{2.19} & 74.70 \stdv{4.01} & \textbf{81.52} \stdv{1.13} \\
BAR           & 70.16 \stdv{0.77} & 71.77 \stdv{0.40} & 71.60 \stdv{0.44} & 72.09 \stdv{0.83} & \textbf{73.36} \stdv{0.97} \\
\bottomrule
\end{tabular}}
\vspace{0.05cm}
\caption{Comparisons between our method and ensembling multiple 1) biased models and 2) biased models in a soft way 3) debiased models. We also report the results of LfF to show performance gain and degradation in each case.}
\label{tab:supp_ensemble}
\end{table*}
\subsection{Ensembling $f_B$}
As mentioned in the main paper, our proposed method is different from simply ensembling multiple biased models. 
We conducted experiments to demonstrate that our proposed method is superior to simply ensembling multiple biased models.
For the experiment, we jointly trained $f_D$ with multiple biased models ($f_{B_1}$, $f_{B_2}$, ... $f_{B_5}$) during the main stage of debiasing. 
Note that $f_{B_i}$ are biased models which are jointly trained with $f_D$ for computing reweighting value $w$, not the pretrained biased models which are denoted as $f^*_{B_i}$ used to filter bias-conflicting samples in our method.
For the reweighting value $w$, we computed the Cross Entropy loss with the averaged softmax probabilities of the five $f_{B_i}$ models which is formulated as 
\begin{equation}
        w(x) = \frac{\mathcal{L_\text{CE}}(\frac{1}{M}\sum^{M}_{i=1} f_{B_i}(x), y)}{\mathcal{L_\text{CE}}(\frac{1}{M}\sum^{M}_{i=1} f_{B_i}(x), y) + \mathcal{L_\text{CE}}(f_D(x), y)}
\end{equation}
where $M$ is set to 5.

\begin{figure*}[t]
    \centering
    \includegraphics[width=0.85\textwidth]{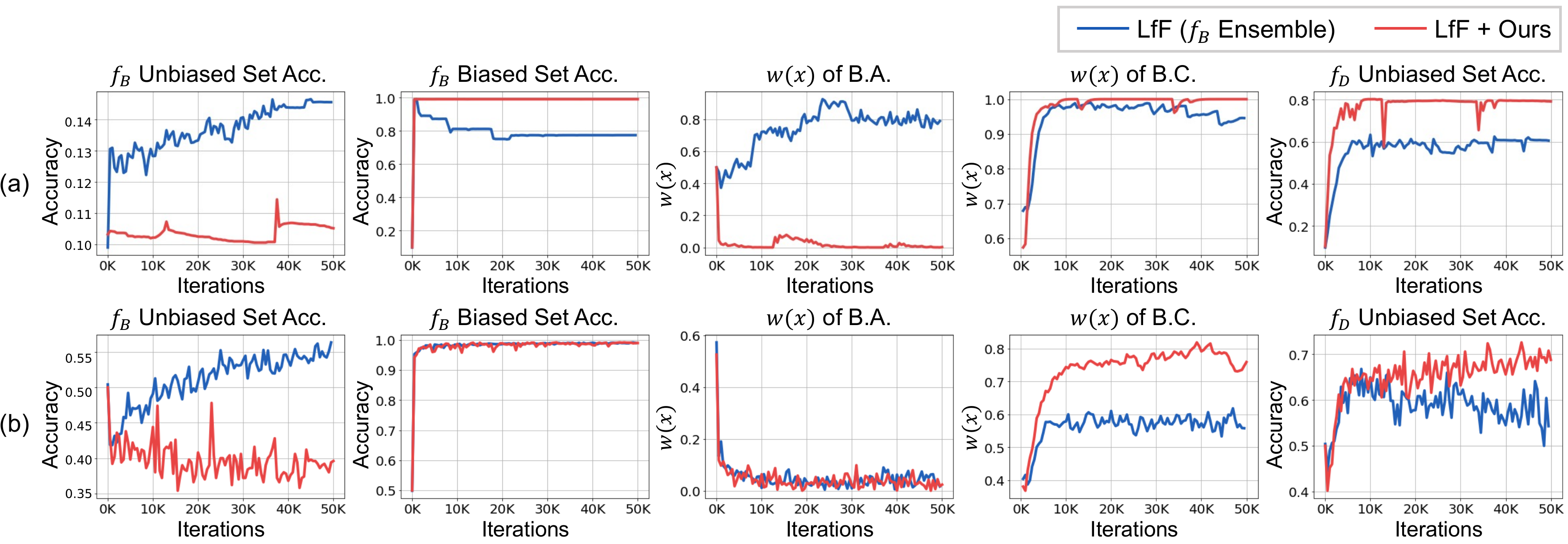}
    \caption{
    Comparison of LfF 1) with ensembling multiple $f_B$ models (blue) and 2) with our method (red). (a) and (b) indicate the results on Colored MNIST and BFFHQ, respectively. Each metric corresponds to the ones used in Fig. 2 of the main paper. For column 1 and column 2, we averaged the values of multiple $f_{B_i}$ for reporting the case of ensembling multiple $f_{B_i}$.
    }
    \vspace{0.2cm} 
    \label{fig:supple-ensemble-graph}
\end{figure*}

Table~\ref{tab:supp_ensemble} shows that our proposed method is superior to ensembling multiple biased models in the main stage of debiasing. 
The main reason is that the bias-conflicting samples are not discarded when training them, so each biased model $f_{B_i}$ gradually learns the intrinsic attribute as training proceeds.
Fig.~\ref{fig:supple-ensemble-graph} supports the point.
Fig.~\ref{fig:supple-ensemble-graph} visualizes each metric used in Fig.~2 of the main paper.
We report the averaged values of $f_{B_i}$ for the accuracies of $f_B$ on the unbiased test set and biased test set. 
We observe that the average accuracy of $f_{B_i}$ models on the unbiased test set increases as training steps proceed even with ensembling them. 

Another variant of our work is to ensemble $f_B$ in a soft way. To be more specific, instead of discarding samples having agreements less than half of the number of BCDs, we can reweight samples by setting $w(x)$ as the ratio of agreements which can be formulated as,

\begin{equation}
        w(x) = \frac{\sum_{i=1}^M{BCD(x; \tau, f_{B_i}^*)}}{M}.
\end{equation}

Table~\ref{tab:supp_ensemble} shows that such a soft-version of our method also improves the debiasing performance of LfF but shows inferior performance compared to our original method. 
However, we want to emphasize that such an approach again shows the importance of further amplifying bias of $f_B$ for improving the debiasing performance of $f_D$ which is the main contribution of our work.

\subsection{Ensembling $f_D$}
One might also question the effectiveness of ensembling $f_D$ in the main stage of debiasing. 
We want to clarify that the main contribution of our paper is revealing that amplifying bias via utilizing predictions of multiple pretrained \emph{biased} models is crucial for improving debiasing performance.
In other words, ensembling multiple \emph{debiased} models is orthogonal to our work. 
In Table~\ref{tab:supp_ensemble}, we also ensemble multiple debiased models of LfF.
We observe that ensembling multiple debiased models in LfF brings performance gain compared to the original training of LfF.
However, applying our sample selection method to LfF without ensembling multiple debiased models still shows superior performance to ensembling multiple debiased models of LfF.
Note that ensembling of debiased models requires large computational cost and memory space for inference, unlike our proposed method.



\begin{figure*}[h!]
    \centering
    \includegraphics[width=0.7\linewidth]{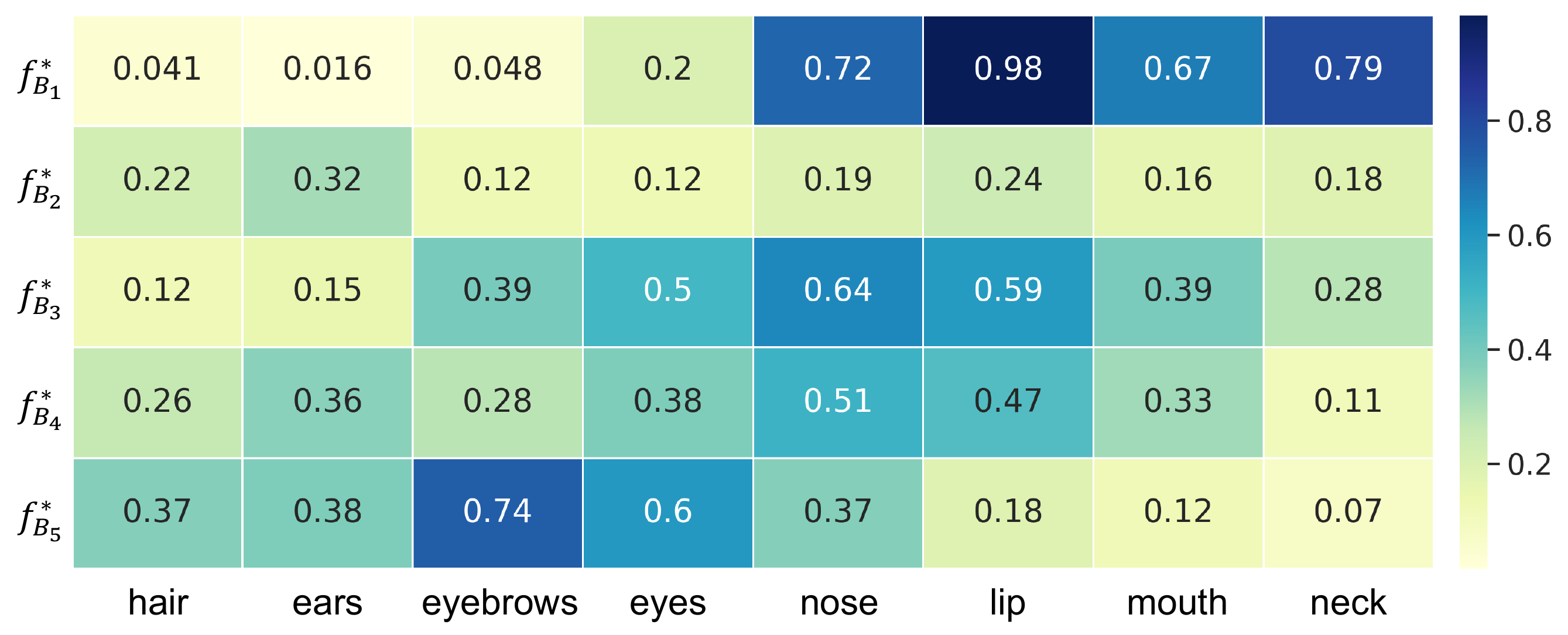}
    \caption{Recall values of overlapped regions between the highlighted pixels and the segmentation map of each facial attribute. The values are different among each $f_{B_i}$, indicating that differently initialized models focus on different facial attributes.}
    \label{fig:gradcam_heatmap}
\end{figure*}

\begin{figure*}[h!]
    \centering
    \includegraphics[width=1.0\textwidth]{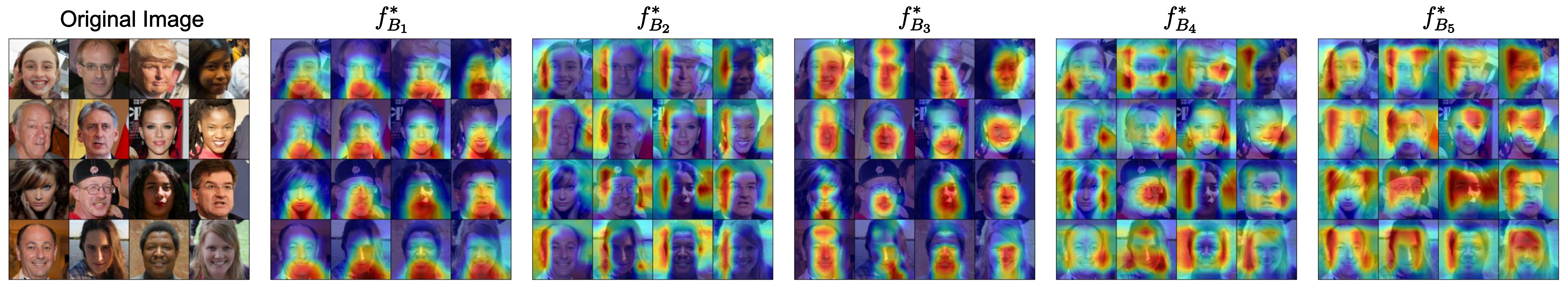}
    \caption{Additional results of GradCAM on each $f_{B}^*$. The first column indicates the original images. Starting from the second column, we visualize GradCAM using each biased model~(\textit{i.e.,} $f_{B_1}^*,\dots,f_{B_5}^*$).}
    \vspace{-0.25cm}
    \label{fig:supple-gradcam}
\end{figure*}

\begin{table*}[h!]
\centering
\resizebox{0.95\linewidth}{!}{
\setlength{\tabcolsep}{1.0em}
\begin{tabular}{c | c c c c c c}
\toprule 
\multirow{4}{*}{Method} & 
\multicolumn{2}{c}{MLP (CMNIST)} &
\multicolumn{2}{c}{ResNet18 (BFFHQ / Dogs\&Cats)}  &
\multicolumn{2}{c}{ResNet18 (BAR)}  \\
\cmidrule(l{2pt}r{2pt}){2-3} \cmidrule(l{2pt}r{2pt}){4-5} \cmidrule(l{2pt}r{2pt}){6-7}
& Training iterations & \# of Params & Training iterations & \# of Params & Training iterations & \# of Params \\
\midrule
LfF & 50K & 0.494M & 50K & 22.355M & 50K & 22.359M \\
LfF + BE & 55K & 0.494M & 55K & 22.355M & 55K & 22.359M \\
\midrule
DisEnt & 50K & 0.495M & 50K & 22.357M & 50K & 22.365M \\
DisEnt + BE & 55K & 0.495M & 55K & 22.357M & 55K & 22.365M\\
\bottomrule
\end{tabular}
}
\caption{The number of training iterations and the number of parameters that need to be loaded in the memory during training. }
\label{table:supp_time_params}
\end{table*}
\begin{table*}[th!]
\small
\begin{center}
\resizebox{\linewidth}{!}{%
\large
\setlength{\tabcolsep}{0.5em}
\def\arraystretch{1.5}%
\begin{tabular}{c|ccc|ccc|ccc|ccc}
\toprule
\multicolumn{1}{c}
{\multirow{2}{*}{Method}}
& \multicolumn{3}{c}{Colored MNIST}
& \multicolumn{3}{c}{BFFHQ}
& \multicolumn{3}{c}{Dogs \& Cats}
& \multicolumn{3}{c}{BAR}
\\ \cmidrule(lr){2-4} \cmidrule(lr){5-7} \cmidrule(lr){8-10} \cmidrule(lr){11-13}

\multicolumn{1}{c}{} & unbiased$\downarrow$ & biased$\uparrow$ & $\text{acc}_{\text{diff}}$$\uparrow$ & unbiased$\downarrow$ & biased$\uparrow$  & $\text{acc}_{\text{diff}}$$\uparrow$ & unbiased$\downarrow$ & biased$\uparrow$  & $\text{acc}_{\text{diff}}$$\uparrow$ & unbiased$\downarrow$ & biased$\uparrow$  & $\text{acc}_{\text{diff}}$$\uparrow$ \\ 
\midrule
$f_B$ of LfF~\cite{nam2020learning} & 15.02 \stdv{1.22} & 88.38 \stdv{14.61} & 73.36 & 55.84 \stdv{1.80} & 99.50 \stdv{0.21} & 43.66 & 36.3 \stdv{3.49} & 96.18 \stdv{0.62} & 59.88 & 69.07 \stdv{0.87} & 98.33 \stdv{0.78} & 29.26 \\
$f_B$ of LfF + BE & \cellcolor{Gray} 13.29 \stdv{1.61}& \cellcolor{Gray}99.10 \stdv{0.05} & \cellcolor{Gray} \textbf{85.81} & \cellcolor{Gray}37.96 \stdv{1.77} & \cellcolor{Gray}98.9 \stdv{0.16} & \cellcolor{Gray} \textbf{60.94} & \cellcolor{Gray}13.42 \stdv{1.42}& \cellcolor{Gray}94.35 \stdv{0.19} & \cellcolor{Gray} \textbf{80.93} & \cellcolor{Gray}53.53 \stdv{2.90}& \cellcolor{Gray}96.00 \stdv{0.73} & \cellcolor{Gray} \textbf{42.47} \\
\bottomrule
\end{tabular}%
}
\caption{Comparisons of $f_B$ on the unbiased test set and the biased test set. Utilizing our method encourages $f_B$ to be further biased in LfF. We use 1\% ratio of bias-conflicting samples for each dataset.
Applying our method on LfF is shaded in grey.
For $\text{acc}_{\text{diff}}$, we compute the difference between the average of biased test set accuracy and that of unbiased test set accuracy.}
\label{tab:supple_bias_unbiased}
\end{center}
\end{table*}

\vspace{-0.3cm}
\begin{table}[h!]
\centering
\resizebox{1.0\linewidth}{!}{
\setlength{\tabcolsep}{1.0em}
\begin{tabular}{c | c c c c}
\toprule 
Method & Colered MNIST & BFFHQ & Dos \& Cats & BAR \\
\midrule
LfF (55K) & 76.96\stdv{3.72} & 70.70 \stdv{2.78} & 75.06 \stdv{5.19} & 71.84\stdv{1.45} \\
LfF + BE & \textbf{81.17} \stdv{0.68} & \textbf{75.08} \stdv{2.29} & \textbf{81.52} \stdv{1.13} & \textbf{73.36} \stdv{0.97} \\
\midrule
DisEnt (55K) & 80.88 \stdv{1.41} & 66.15 \stdv{1.04} & 67.94 \stdv{2.42} & 70.12 \stdv{1.37} \\
DisEnt + BE & \textbf{82.11} \stdv{0.54} & \textbf{73.48} \stdv{2.12} & \textbf{80.74} \stdv{2.80} & \textbf{73.29} \stdv{0.41}\\
\bottomrule
\end{tabular}
}
\caption{Comparisons of unbiased test set accuracies of reweighting-based methods~(\textit{i.e.,} LfF, DisEnt) trained for 55K iterations and our method. Simply training with additional 5K iterations, the number of training steps used in our method, on existing reweighting-based debiasing approaches does not bring performance gain.}
\label{table:supp_55k}
\vspace{-0.3cm}
\end{table}

\section{Multiple $f_{B}^*$ Capturing Diverse Visual Attributes}
\label{supple:gradcam}
In the main paper, we showed the results of GradCAM visualization of each $f_{B}^*$ which demonstrate that each biased model utilizes different facial attributes for making biased predictions. 
This section supports such a fact by showing the quantitative measurement of how each biased model focuses on different parts of images. 
By using the GradCAM visualization of each image, we obtained pixels of an image whose the values of GradCAM are higher than a certain threshold.
We name such pixels as `highlighted pixels'.
Since the scale of GradCAM is between 0 to 1, we chose the threshold as 0.5.
Then, we utilized a segmentation mask~\cite{or2020lifespan} of FFHQ~\cite{stylegan} including semantic classes such as eyes, lips, hair, and so on.
We compute the recall value, $S$, of the overlapped region between the highlighted pixels and the segmentation map of each facial attribute which can be formulated as, 
\begin{equation}
    S = \frac{A_{overlap}}{A_{mask}}
\end{equation}
where $A_{overlap}$ and $A_{mask}$ indicate the area of the overlapped region and that of the segmentation mask of each facial attribute, respectively. 
In Fig.~\ref{fig:gradcam_heatmap}, we observe that $S$ values of the facial attributes are different in each $f_{B_i}^*$.
For example, $S$ values of lip and mouth are high for $f_{B_1}^*$ while those of eye and eyebrows are high for $f_{B_5}^*$. 
This again demonstrates that differently initialized biased models focus on different visual attributes. 
Note that among 19 facial attributes in the segmentation mask dataset, we selectively chose the ones which may contribute to making biased predictions towards gender.
While ears and neck may not contribute to predicting the gender, they are located near to the facial attributes which may contribute to making biased predictions (\textit{e.g.,} ears are close to hair and neck is close to the lip).
Thus, we included such visual attributes in our analysis. 

As done in the main paper, we show additional results of GradCAM visualization in Fig.~\ref{fig:supple-gradcam}. 
We again observe that each $f_{B}^*$ focuses on different part of image for making biased predictions. 
\section{Analysis on Training Steps and Memory Space}

\noindent\textbf{Training Steps. \enskip}
We compare the number of iterations for training models with and without our approach in Table~\ref{table:supp_time_params}.
Since we train each biased model $f_B^*$ with a small number of iterations (1K), the additional computational cost of our approach is marginal.  
Our data selection module only necessitates additional 5K iterations while iteratively pretraining the five biased models. 
Also, Table~\ref{table:supp_55k} demonstrates that simply using additional 5K training steps on existing reweighting-based debiasing approaches does not bring performance gain. 
In other words, it is necessary to utilize additional small number of training steps in an adequate manner. 

\noindent\textbf{Memory Space. \enskip}
Generally, ensemble methods need multiple model checkpoints to be stored in memory to make multiple predictions at the \emph{inference time}.
However, unlike such ensemble methods, our approach only needs to assign pseudo labels on the \emph{training samples}, so we do not have to save the model checkpoints for the purpose of using them at the \emph{test phase}.
Without saving the multiple model checkpoints in the memory space, we obtain the predictions of a biased model on the training samples, and re-initialize the next biased model with the same memory.
Table~\ref{table:supp_time_params} demonstrates that our method does not require additional number of parameters for the existing reweighting-based debiasing methods. 
We explain the detailed procedure with a pseudo code in the section named "Pseudo-code of Our Training Procedure" in this Supplementary regarding such an issue.

\section{Amplified Bias via Bias Amplified Dataset}
\label{supple:amplified-bias}

Table~\ref{tab:supple_bias_unbiased} shows the image classification accuracy of $f_B$ on the biased and unbiased test set. 
We compare LfF~\cite{nam2020learning} 1) without our method and 2) with our method using the test set accuracy at the last epoch in Table~\ref{tab:supple_bias_unbiased}. 
As mentioned in the main paper, a biased model achieves 1) low accuracy on the unbiased test set and 2) high accuracy on the biased test set.
In other words, the difference between the accuracy of the biased test set and that of the unbiased test set, $\text{acc}_{\text{diff}}$ in Table~\ref{tab:supple_bias_unbiased}, is high for the biased models.
This is due to the fact that the biased model heavily relies on the bias attributes while failing to learn the intrinsic attributes. 

As shown in Table~\ref{tab:supple_bias_unbiased}, using our biased sample selection module significantly increases the $\text{acc}_{\text{diff}}$ indicating that utilizing our method further amplifies bias for $f_B$. 
To be more specific, applying our method achieves low unbiased test set accuracy and high biased test set accuracy in Colored MNIST.
Meanwhile, we observe a minimal performance drop in the biased test set with other datasets.
However, considering the significant performance drop on the unbiased test set and high $\text{acc}_{\text{diff}}$, such a drop in biased test set accuracy is negligible.
The result shows that the debiasing performance gain of our method is due to the amplified bias of $f_B$.
It again demonstrates the fact that two factors are crucial for amplifying bias: 1) improving the biased test set accuracy of $f_B$ and 2) degrading the unbiased test set accuracy of $f_B$ (\textit{i.e.,} not learning intrinsic attributes).

\begin{figure*}[h!]
    \centering
    \includegraphics[width=0.9\linewidth]{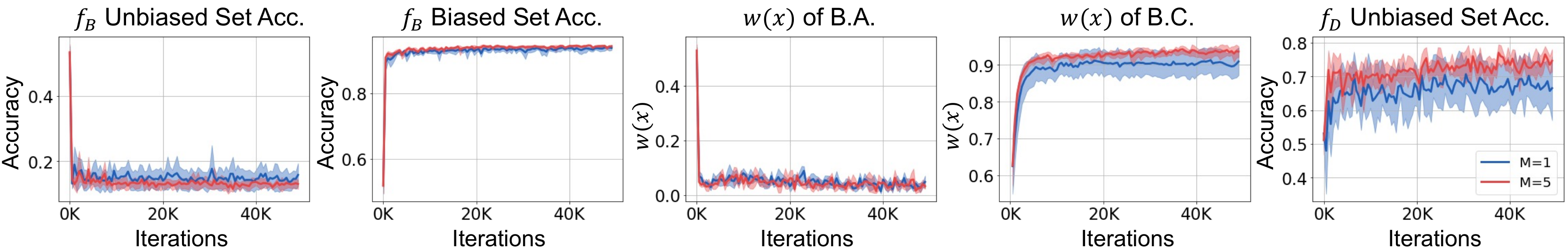}
    \vspace{-0.2cm}
    \caption{
    Comparison of models trained with 1) a single BCD ($M=1$, blue) and 2) multiple BCDs ($M=5$, red) on Dogs \& Cats. 
    The shaded areas show the variations of the results on five different independent trials.
    Each metric corresponds to the ones used in Fig.~2 of the main paper.}
    \vspace{-0.3cm}
    \label{fig:ens_ablation}
\end{figure*}
\vspace{0.2cm}
\begin{table*}[t!]
\centering
\resizebox{0.85\linewidth}{!}{
\setlength{\tabcolsep}{1.0em}
\begin{tabular}{c c c c c c c c}
\toprule

\multirow{4}{*}{Dataset} & 
\multicolumn{3}{c}{Unbiased test set accuracy} &
\multicolumn{4}{c}{Remaining samples in $\mathcal{D}_{A}$}                  \\
\cmidrule(l{2pt}r{2pt}){2-4} \cmidrule(l{2pt}r{2pt}){5-8}
 & \multirow{2.5}{*}{LfF} & \multirow{2.5}{*}{$\mathcal{N}_a$=$5$} & \multirow{2.5}{*}{$\mathcal{N}_a$=$3$} & \multicolumn{2}{c}{\# of B.A.(\%)$\uparrow$} & \multicolumn{2}{c}{\# of B.C.(\%)$\downarrow$} \\
 \cmidrule(l{2pt}r{2pt}){5-6} \cmidrule(l{2pt}r{2pt}){7-8}
 & & & & $\mathcal{N}_a$=$5$ & $\mathcal{N}_a$=$3$ & $\mathcal{N}_a$=$5$ & $\mathcal{N}_a$=$3$ \\
\midrule
Colored MNIST & 76.81\stdv{4.56}	&	80.51\stdv{0.88} & \textbf{81.17}\stdv{0.68} & 96.80\stdv{1.06} & \textbf{99.96}\stdv{0.03} & \textbf{0.00}\stdv{0.00} & 1.42\stdv{0.78} \\

BFFHQ         & 69.24\stdv{2.07}	&	70.00\stdv{1.44} & \textbf{75.08}\stdv{2.29} & 57.48\stdv{7.36} & \textbf{92.22}\stdv{0.26} & \textbf{2.29}\stdv{1.08}  & 24.37\stdv{3.07} \\

Dogs \& Cats  & 71.72\stdv{4.56}	&	76.86\stdv{2.11} & \textbf{81.52}\stdv{1.13} & 57.84\stdv{1.52} & \textbf{88.60}\stdv{1.02} & \textbf{1.00}\stdv{0.56}  & 9.50\stdv{3.75} \\

BAR           & 70.16\stdv{0.77}	&	71.84\stdv{1.22} & \textbf{73.36}\stdv{0.97} & 63.79\stdv{0.81} & \textbf{98.39}\stdv{0.19} & \textbf{5.71}\stdv{3.19}  & 51.42\stdv{5.34} \\

\bottomrule
\end{tabular}}
\vspace{-0.2cm}
\caption{Unbiased test set accuracies and remaining samples in $\mathcal{D_{A}}$ depending on the number of agreement votes for discarding bias-conflicting samples. Using $\mathcal{N}_a$=5 rather aggravates the debiasing performance since it discards a significant number of bias-aligned samples which is important for constructing a bias-amplified dataset $\mathcal{D}_A$.}
\vspace{-0.4cm}
\label{tab:ablation_voting_five}
\end{table*}

\section{Performance Variation of Single/Multiple BCDs}
\label{supple:stable-multiple}
As mentioned in the main paper, utilizing multiple BCDs reduces the performance variations compared to using a single BCD.
Fig.~\ref{fig:ens_ablation} demonstrates the fact by visualizing the metrics used in Fig.~2 of the main paper with Dogs \& Cats dataset. 
The shaded regions indicate the performance variation, the differences between five independent trials on a given metric. 
We observe that utilizing multiple BCDs (red) shows smaller variations in terms of $w(x)$ of bias-conflicting samples and the unbiased test set accuracy of $f_D$.
Again, this is due to the fact that leveraging multiple numbers of $f_B^{*}$ enables to consider multiple visual attributes, better capturing the bias attribute compared to using a single $f_B^{*}$.

\section{Biased Sample Selection with Strict Agreements}
\label{supple:be-strict}

There are two factors crucial for constructing a bias-amplified dataset $\mathcal{D}_A$: 1) discarding a sufficient number of bias-conflicting samples and 2) maintaining a reasonable number of bias-aligned ones. 
We further demonstrate such a point by designing a strict version of the sample selection module in Table~\ref{tab:ablation_voting_five}; we regard a sample as a bias-conflicting one only if all five BCDs consider it as a bias-conflicting sample.
$\mathcal{N}_{a}$ in Table~\ref{tab:ablation_voting_five} indicates the number of agreement votes required for deciding a sample as a bias-conflicting one. 
We add our sample selection method on LfF~\cite{nam2020learning} with $\mathcal{N}_{a}$ set to three and five, and compare them with the original LfF. 
Note that the number of BCDs, $M$, is fixed to five.

Utilizing our biased sample selection method with $\mathcal{N}_{a}$=5 outperforms the original LfF.
However, we observe that utilizing the selection method with $\mathcal{N}_{a}$=3 is superior to the method with $\mathcal{N}_{a}$=5. 
The main reason is due to the number of bias-aligned and bias-conflicting samples. 
Similar to Table~4 of the main paper, we report the remaining samples of bias-aligned samples and bias-conflicting samples in $\mathcal{D}_A$ by showing the adjusted ratio compared to each group in $\mathcal{D}$.
We observe that utilizing our method with $\mathcal{N}_{a}$=5 discards the bias-conflicting ones significantly while also filtering out a considerable number of the bias-aligned ones.
For example, by utilizing $\mathcal{N}_{a}$=5 in BFFHQ, there exist only 2.29\% of bias-conflicting samples in $\mathcal{D}_A$ compared to $\mathcal{D}$.
However, only 57.48\% of bias-aligned samples are remained in $\mathcal{D}_A$, limiting $f_B$ to be further biased. 

\vspace{0.4cm}
\begin{table}[t!]
\centering
\resizebox{1.0\linewidth}{!}{
\setlength{\tabcolsep}{0.5em}
\begin{tabular}{ c | c c c c}

\toprule 

Loss & Colored MNIST & BFFHQ & Dogs \& Cats & BAR \\
\midrule
CE & 66.90 \stdv{4.84} & 70.90 \stdv{0.81} & 71.84 \stdv{3.46} & 72.48 \stdv{1.24} \\
GCE & \textbf{81.17} \stdv{0.68} & \textbf{75.08} \stdv{2.29} & \textbf{81.52} \stdv{1.13}& \textbf{73.36} \stdv{0.97} \\
\bottomrule
\end{tabular}}
\vspace{-0.2cm}
\caption{Unbiased test set accuracies of LfF with our sample selection module. The first and second row represent the result of the method of using CE and GCE loss for training $f^*_B$, respectively. We use 1\% ratio of bias-conflicting samples for each dataset.}
\vspace{-0.5cm}
\label{tab:supp_ce_gce}
\end{table}




\vspace{-0.2cm}
\section{Training $f^*_B$ with CE vs GCE}
As mentioned in the main paper, we pretrain multiple $f^*_{B_i}$ for a small number of iterations by using the property that the bias is easy to learn. 
Although the bias attribute is relatively easy to learn compared to the intrinsic attribute, we need to utilize the GCE loss rather than the CE loss. 
We compared the debiasing performance between using CE loss and GCE loss for pretraining multiple $f^*_{B_i}$ in our method. 
We observe that training multiple $f^*_{B_i}$ with GCE loss is superior to training it with CE loss.
This is because GCE loss further amplifies bias compared to the original CE loss by emphasizing the easy-to-learn samples (\textit{i.e.,} bias-aligned samples).
GCE loss has been widely used in the previous debiasing studies~\cite{nam2020learning,disentangled} for emphasizing bias-aligned samples, resulting a model to be overfitted to a bias overall. 

\vspace{-0.25cm}
\section{Further Implementation Details}
\label{supple:implementation}
\noindent \textbf{Experimental settings. \enskip}
We set the learning rate as 0.01, 0.0001, 0.0001, and 0.00001 for Colored MNIST, BFFHQ, Dogs \& Cats, and BAR, respectively. 
The image sizes are 28$\times$28 for Colored MNIST and 224$\times$224 for the rest of the datasets. 
For Colored MNIST, we do not apply additional data augmentation techniques.
For BFFHQ, Dogs \& Cats, and BAR, we apply random crop and horizontal flip transformations. 
Also, images are normalized along each channel (3, H, W) with the mean of (0.4914, 0.4822, 0.4465) and standard deviation of (0.2023, 0.1994, 0.2010). 
We conducted all experiments using a single V100 GPU. 
We used the default values of hyperparameters reported in the original paper for the baseline models.
Applying our method on LfF~\cite{nam2020learning} requires approximately up to 0.8GB of memory for Colored MNIST and 4.3GB of memory for the rest of the datasets.
\newline 

\noindent \textbf{Applying Our Method on Reweighting-based Methods.\enskip}
As aforementioned in the main paper, when applying our method to the existing reweighting-based approaches (\textit{i.e.,} LfF~\cite{nam2020learning} and DisEnt~\cite{disentangled}), we do not modify the training procedure of the debiased model $f_{D}$.
For both methods, we train $f_{B}$ using the bias-amplified dataset $\mathcal{D}_A$.
DisEnt, unlike LfF, additionally utilizes feature-level data augmentation during training.
When we apply our selection method to DisEnt, we use $\mathcal{D}_A$ for training the parameters which are updated with respect to the feature vectors of bias attributes.
$f_B$ of our paper corresponds to $E_b$ and $C_b$ in the original paper of DisEnt~\cite{disentangled}.

\begin{figure*}[t!]
    \centering
    \includegraphics[width=0.85\linewidth]{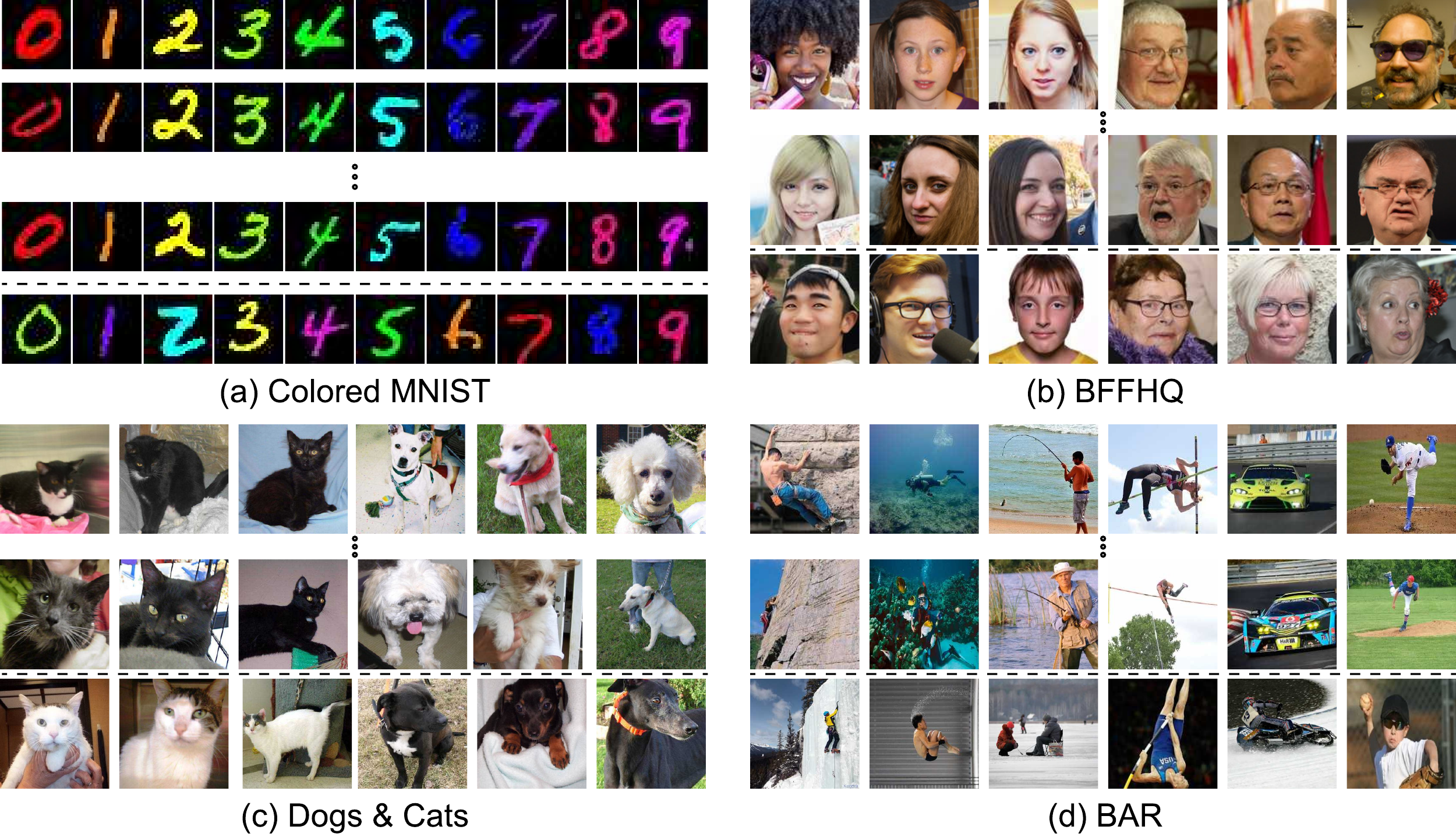}
    \vspace{+0.5cm}
    \caption{Illustrative example of the datasets used in our work.
    Each column in Colored MNIST and BAR dataset indicates each class, while the group of three columns in BFFHQ and Dogs \& Cats indicate each class. The images above the dotted line are the bias-aligned samples, while the ones below indicate the bias-conflicting samples. 
    }
    \vspace{+1cm}
    \label{fig:supple-dataset}
\end{figure*}

\begin{table*}[h!]
\centering
\resizebox{0.95\linewidth}{!}{
\setlength{\tabcolsep}{0.3em}
\def\arraystretch{1.2}%
\begin{tabular}{c | c | c c c c | c c c c}
\toprule 

\# of B.A.          & 100\%            & 100\%             & 100\%            & 100\%            & 100\%                      & 20\%                      & 40\%             & 60\%             & 80\%    \\
\# of B.C.          & 100\%            & 80\%              & 60\%             & 40\%             & 20\%                       & 20\%                      & 20\%             & 20\%             & 20\%    \\
\midrule
Colored MNIST       & 58.48\stdv{2.78} &  66.47\stdv{2.86} & 73.87\stdv{2.55} & 77.87\stdv{3.30} & \textbf{81.58}\stdv{0.68}  & 63.21\stdv{2.86}          & 73.86\stdv{2.31} & 75.75\stdv{5.10} & \textbf{79.82}\stdv{0.65} \\
BFFHQ               & 62.10\stdv{0.79} & 69.88\stdv{0.83}  & 73.96\stdv{2.44} & 77.60\stdv{1.72}  & \textbf{79.36}\stdv{1.43}  & 69.12\stdv{1.12}          & 72.20\stdv{0.88}  & 75.88\stdv{2.57} & \textbf{78.52}\stdv{1.19} \\
Dogs\&Cats          & 53.10\stdv{4.80}  & 61.18\stdv{3.70}  & 71.00\stdv{2.85} & 75.38\stdv{3.27} & \textbf{79.22}\stdv{2.83}  & 60.78\stdv{3.53}          & 60.58\stdv{2.18} & 63.86\stdv{4.49} & \textbf{71.80}\stdv{2.04}     \\

\bottomrule
\end{tabular}}
\vspace{0.1cm}
\caption{
Unbiased test set accuracies with adjusted number of bias-aligned samples~(\# of B.A.) and that of bias-conflicting ones~(\# of B.C.) in the bias-amplified dataset $\mathcal{D}_A$ utilized for training $f_B$ of LfF with CE loss.
The values in the first two rows represent the ratio of samples in $\mathcal{D}_A$ compared to $\mathcal{D}$. 
}
\vspace{1cm}
\label{tab:supp_ablation_align_conflict}
\end{table*}

\begin{table*}[h!]
\vspace{0.15cm}
\begin{center}
\label{table:headings}
\resizebox{0.9\linewidth}{!}{
\setlength{\tabcolsep}{1.3em}
\begin{tabular}{c|c|ccccc}
\toprule 
Method & LfF & \multicolumn{5}{c}{LfF + BE} \\
\midrule
Threshold & - & 0.9	& 0.95  & 0.98 & 0.99 & 0.999 \\
\midrule
Colored MNIST & 76.81\stdv{4.56} & 80.65\stdv{1.88} & 80.49\stdv{2.16} & 80.41\stdv{1.51}  & \textbf{81.17}\stdv{0.68} &    81.06\stdv{0.86} \\

BFFHQ         & 69.24\stdv{2.07} & 73.16\stdv{1.90} & 74.48\stdv{0.89} & 74.28\stdv{1.78}  & \textbf{75.08}\stdv{2.29} &	72.32\stdv{1.57} \\

Dogs \& Cats  & 71.72\stdv{4.56} & 80.60\stdv{2.87} & 77.66\stdv{3.37} & 80.36\stdv{2.69}  & \textbf{81.52}\stdv{1.13} &	77.90\stdv{1.50} \\
    
BAR           & 70.16\stdv{0.77} & 71.53\stdv{1.98} & 72.23\stdv{0.45} & 73.00\stdv{1.52}  & \textbf{73.36}\stdv{0.97} &	71.28\stdv{0.88} \\



    

\bottomrule
\end{tabular}}
\vspace{-0.3cm}
\caption{
    Unbiased test set accuracies evaluated with a wide range of confidence threshold $\tau$ of our method.
    }
\label{tab:supp_ablation_threshold}
\vspace{1cm}
\end{center}
\end{table*}


\section{Additional Experimental Results}
We additionally report the results which we intentionally omitted in the main paper due to the page limit. 
Table~\ref{tab:supp_ablation_align_conflict} shows the tendency of debiasing performance with additional cases of adjusted numbers of bias-aligned and bias-conflicting samples (Table~2 of the main paper). 
We also included the standard deviation of each result. 
Table~\ref{tab:supp_ablation_threshold} includes the standard deviations of results demonstrating the robustness of our proposed method to a wide range of confidence threshold $\tau$ (Table~5 of the main paper).

\vspace{-0.1cm}
\section{Further Details on Datasets}
\label{supple:dataset}
Fig.~\ref{fig:supple-dataset} shows the dataset used in our work.

\noindent \textbf{Colored MNIST. \enskip} 
We modify MNIST dataset~\cite{mnist} and set the color as the bias attribute. 
Each one of the ten digits is highly correlated with a certain color (\textit{e.g.,} red for 0). We inject the color into the foreground of each image. 
For Colored MNIST, we conduct experiments with the dataset used in Lee~\textit{et al.}~\shortcite{disentangled}.
For each ratio of bias-conflicting samples, the number of bias-aligned samples and bias-conflicting samples are as follows: ($54751$, $249$)-$0.5\%$, ($54509$, $491$)-$1\%$, ($54014$, $986$)-$2\%$, and ($52551$, $2449$)-$5\%$.

\noindent \textbf{BFFHQ. \enskip} 
We used BFFHQ which was first introduced in Kim~\textit{et al.}~\shortcite{biaswap}. 
BFFHQ sets the age (\textit{i.e.,} old and young) as the intrinsic attribute and the gender as the bias attribute. 
Bias-aligned samples are young (\textit{i.e.,} age ranging from 10 to 29) females and old (\textit{i.e.,} age ranging from 40 to 59) males. 
For each ratio of bias-conflicting sample, the number of bias-aligned samples and bias-conflicting samples are as follows: ($19104$, $96$)-$0.5\%$, ($19008$, $192$)-$1\%$, ($18816$, $384$)-$2\%$, and ($18240$, $960$)-$5\%$.
Following the work of Lee~\textit{et al.}~\shortcite{disentangled}, we exclude the bias-aligned samples from the unbiased test set since it only includes two target classes.

\noindent \textbf{Dogs \& Cats. \enskip}
Dogs \& Cats~\footnote{\url{https://www.kaggle.com/c/dogs-vs-cats-redux-kernels-edition}} was first used in Kim et al.~\shortcite{LNL}.
However, we find that the test sets used in the work do not have publicly available ground truths. 
Therefore, since the original train set has ground truth labels, we split the original train set into train, valid, and test sets.
For each ratio of bias-conflicting sample, the number of bias-aligned samples and bias-conflicting samples are as follows: ($8037$, $80$)-$1\%$ and ($8037$, $452$)-$5\%$.
We did not include the bias-aligned samples in the unbiased test set. 

\noindent \textbf{BAR. \enskip} 
In the work of Nam~\textit{et al.}~\shortcite{nam2020learning}, the original BAR dataset does not include bias-conflicting samples in the training set.
However, for the consistent experimental setting and for the evaluation under various bias severities, we build two sets of BAR dataset with different ratios of bias-conflicting samples. 
For each ratio of bias-conflicting sample, the number of bias-aligned samples and bias-conflicting samples are as follows: ($1761$, $14$)-$1\%$ and ($1761$, $85$)-$5\%$.
We did not include the bias-aligned samples in the unbiased test set.

\begin{table*}[b]
\small
\begin{center}
\resizebox{\linewidth}{!}{
\large
\setlength{\tabcolsep}{1.5em}
\def\arraystretch{1}%
\begin{tabular}{cc|cccc|cc}
\toprule
\multicolumn{2}{c}
{\multirow{2}{*}{Method}}
& \multicolumn{4}{c}{Colored MNIST}
& \multicolumn{2}{c}{Dogs \& Cats}
\\ \cmidrule(lr){3-6}\cmidrule(lr){7-8}
\multicolumn{2}{c}{}
& 0.5\% & 1.0\% & 2.0\% & 5.0\% & 1.0\% & 5.0\% \\
\midrule
Vanilla~\cite{He2015resnet} & \textcolor{black}{\boldxmark} \textcolor{black}{\boldxmark} 
& 34.75\stdv{1.68}  & 51.14\stdv{3.12}  & 65.72\stdv{3.74}  & 82.82\stdv{0.63}   & 48.06\stdv{7.09}  & 69.88\stdv{1.73}  \\
HEX~\cite{wang2018hex} & \textcolor{black}{\boldxmark} \textcolor{black}{\boldcheckmark} 
& 42.25\stdv{1.83}  & 47.02\stdv{15.08}  & 72.82\stdv{1.03}  & 85.50\stdv{1.94}   & 46.76\stdv{5.3}  & 72.60\stdv{3.01}  \\
LNL~\cite{LNL} & \textcolor{black}{\boldcheckmark} \textcolor{black}{\boldcheckmark} 
& 36.29\stdv{0.77}  & 49.48\stdv{5.29}  & 63.30\stdv{2.73}  & 81.30\stdv{0.80}   & 50.90\stdv{4.62}  & 73.96\stdv{1.86}  \\
EnD~\cite{EnD} & \textcolor{black}{\boldcheckmark} \textcolor{black}{\boldcheckmark} 
& 35.33\stdv{1.23}  & 48.97\stdv{11.54}  & 67.01\stdv{2.81}  & 82.09\stdv{2.20}   & 48.56\stdv{5.69}  & 68.24\stdv{2.24}  \\
ReBias~\cite{bahng2019rebias} & \textcolor{black}{\boldxmark} \textcolor{black}{\boldcheckmark} 
& 60.86\stdv{1.96}  & \textbf{82.78}\stdv{1.36}  & \textbf{92.00}\stdv{0.47}  & \textbf{96.45}\stdv{0.17}  & 48.70\stdv{6.05}  & 65.74\stdv{0.51}  \\
LfF~\cite{nam2020learning} & \textcolor{black}{\boldxmark} \textcolor{black}{\boldxmark} 
& 63.55\stdv{6.97}  & 76.81\stdv{4.56}  & 84.18\stdv{1.15}  & 89.65\stdv{0.49}   & 71.72\stdv{4.56}  & 84.32\stdv{1.87}  \\
DisEnt~\cite{disentangled} & \textcolor{black}{\boldxmark} \textcolor{black}{\boldxmark} 
& 68.49\stdv{3.36}  & 79.99\stdv{2.15}  & 84.09\stdv{1.46}  & 89.91\stdv{0.55}   & 65.74\stdv{3.31}  & 81.58\stdv{2.44}  \\
\midrule
\multirow{2}{*}{LfF + BE} & \multirow{2}{*}{\textcolor{black}{\boldxmark} \textcolor{black}{\boldxmark}} 
& 69.70\stdv{4.10} & 81.17\stdv{0.68} & 85.20\stdv{0.85} & 90.04\stdv{0.18} & \textbf{81.52}\stdv{1.13} & \textbf{88.60}\stdv{1.79} \\
&  & \cellcolor{Gray} (+ 6.15) & \cellcolor{Gray} (+ 4.36) & \cellcolor{Gray} (+ 1.02) & \cellcolor{Gray} (+ 0.39) & \cellcolor{Gray} (+ 9.80) & \cellcolor{Gray} (+ 4.28) \\
\multirow{2}{*}{DisEnt + BE} & \multirow{2}{*}{\textcolor{black}{\boldxmark} \textcolor{black}{\boldxmark}} 
& \textbf{71.34}\stdv{1.30} & 82.11\stdv{0.54} & 84.66\stdv{1.72} & 90.15\stdv{0.48} & 80.74\stdv{2.80} & 86.84\stdv{0.77} \\
&  & \cellcolor{Gray} \cellcolor{Gray} (+ 2.85) & \cellcolor{Gray} (+ 2.12) & \cellcolor{Gray} (+ 0.57) & \cellcolor{Gray} (+ 0.24) & \cellcolor{Gray} (+ 15.00) & \cellcolor{Gray} (+ 5.26) \\

\bottomrule
\end{tabular}%
}
\caption{Image classification accuracies on unbiased test sets of Colored MNIST and Dogs \& Cats with varying ratios of bias-conflicting samples.
The \textit{cross} and \textit{check} represent whether each model 1) uses bias labels during training and 2) requires predefined bias type.
For LfF and DisEnt, the performance gains are shaded in grey.
Best performing results are marked in bold.}
\label{tab:supp_best_test_v1}
\end{center}
\vspace{1cm}
\end{table*}

\begin{table*}[b]
\small
\begin{center}
\resizebox{\linewidth}{!}{
\large
\setlength{\tabcolsep}{1.0em}
\def\arraystretch{1}%
\begin{tabular}{cc|cccc|cc}
\toprule
\multicolumn{2}{c}
{\multirow{2}{*}{Method}}
& \multicolumn{4}{c}{BFFHQ}
& \multicolumn{2}{c}{BAR}
\\ \cmidrule(lr){3-6}\cmidrule(lr){7-8}
\multicolumn{2}{c}{}
& 0.5\% & 1.0\% & 2.0\% & 5.0\% & 1.0\% & 5.0\% \\
\midrule
Vanilla~\cite{He2015resnet} & \textcolor{black}{\boldxmark} \textcolor{black}{\boldxmark} 
& 55.64\stdv{0.44}  & 60.96\stdv{1.00}  & 69.00\stdv{0.50}  & 82.88\stdv{0.49}  & 70.55\stdv{0.87}  & 82.53\stdv{1.08} \\
HEX~\cite{wang2018hex} & \textcolor{black}{\boldxmark} \textcolor{black}{\boldcheckmark} 
& 56.96\stdv{0.62}  & 62.32\stdv{1.21}  & 70.72\stdv{0.89}  & 83.40\stdv{0.34}  & 70.48\stdv{1.74}  & 81.20\stdv{0.68} \\
LNL~\cite{LNL} & \textcolor{black}{\boldcheckmark} \textcolor{black}{\boldcheckmark} 
& 56.88\stdv{1.13}  & 62.64\stdv{0.99}  & 69.80\stdv{1.03}  & 83.08\stdv{0.93}  & 70.65\stdv{1.28}  & 82.43\stdv{1.25} \\
EnD~\cite{EnD} & \textcolor{black}{\boldcheckmark} \textcolor{black}{\boldcheckmark} 
& 55.96\stdv{0.91}  & 60.88\stdv{1.17}  & 69.72\stdv{1.14}  & 82.88\stdv{0.74}  & 71.07\stdv{2.03}  & 82.78\stdv{0.30} \\
ReBias~\cite{bahng2019rebias} & \textcolor{black}{\boldxmark} \textcolor{black}{\boldcheckmark} 
& 55.76\stdv{1.50}  & 60.68\stdv{1.24}  & 69.60\stdv{1.33}  & 82.64\stdv{0.64}  & 73.04\stdv{1.04}  & 83.90\stdv{0.82} \\
LfF~\cite{nam2020learning} & \textcolor{black}{\boldxmark} \textcolor{black}{\boldxmark} 
& 65.19\stdv{3.23}  & 69.24\stdv{2.07}  & 73.08\stdv{2.70}  & 79.80\stdv{1.09} & 70.16\stdv{0.77}  & 82.95\stdv{0.27} \\
DisEnt~\cite{disentangled} & \textcolor{black}{\boldxmark} \textcolor{black}{\boldxmark} 
& 62.08\stdv{3.89}  & 66.00\stdv{1.33}  & 69.92\stdv{2.72}  & 80.68\stdv{0.25}  & 70.33\stdv{0.19}  & 83.13\stdv{0.46} \\
\midrule
\multirow{2}{*}{LfF + BE} & \multirow{2}{*}{\textcolor{black}{\boldxmark} \textcolor{black}{\boldxmark}} 
& 67.36\stdv{3.10} & \textbf{75.08}\stdv{2.29} & \textbf{80.32}\stdv{2.07} & \textbf{85.48}\stdv{2.88} & \textbf{73.36}\stdv{0.97} & 83.87\stdv{0.82} \\
&  & \cellcolor{Gray} (+ 2.17) & \cellcolor{Gray} (+ 5.84) & \cellcolor{Gray} (+ 7.24) & \cellcolor{Gray} (+ 5.68) & \cellcolor{Gray} (+ 3.20) & \cellcolor{Gray} (+ 0.92) \\
\multirow{2}{*}{DisEnt + BE} & \multirow{2}{*}{\textcolor{black}{\boldxmark} \textcolor{black}{\boldxmark}} 
& \textbf{67.56}\stdv{2.11} & 73.48\stdv{2.12} & 79.48\stdv{1.80} & 84.84\stdv{2.11} & 73.29\stdv{0.41} & \textbf{84.96}\stdv{0.69} \\
&  & \cellcolor{Gray} (+ 5.48) & \cellcolor{Gray} (+ 7.48) & \cellcolor{Gray} (+ 9.56) & \cellcolor{Gray} (+ 4.16) & \cellcolor{Gray} (+ 2.96) & \cellcolor{Gray} (+ 1.83) \\

\bottomrule
\end{tabular}%
}

\caption{Image classification accuracies on unbiased test sets of BFFHQ and BAR with varying ratios of bias-conflicting samples.
The \textit{cross} and \textit{check} represent whether each model 1) uses bias labels during training and 2) requires predefined bias type.
For LfF and DisEnt, the performance gains are shaded in grey.
Best performing results are marked in bold.}
\label{tab:supp_best_test_v2}
\end{center}
\end{table*}

\vspace{-0.2cm}
\section{Limitation}
\label{supple:discussion}

One limitation of our work is that there still exist a small number of bias-conflicting samples in $\mathcal{D}_A$.
Ideally, a model trained with bias-aligned samples only would be perfectly biased towards the bias attribute.
Since we do not use explicit bias labels, we could not discard all bias-conflicting samples when constructing $\mathcal{D}_A$ for training $f_B$.
In the meanwhile, we filtered out a sufficient number of bias-conflicting samples, improving the debiasing performance of $f_D$ which has been demonstrated throughout the paper.

\section{Relation to Spatial Heterogeneity}
Studies in spatial heterogeneity~\cite{spatial-hetero} has a similar characteristic in that they also filter out subsets of a dataset using error distributions.
However, the main difference is that the debiasing task assumes an imbalanced number of bias-aligned and bias-conflicting ones while studies in spatial heterogeneity do not assume such an imbalanced setting.
Additionally, recent debiasing studies assume that prior knowledge on whether a given sample is a bias-conflicting one or a bias-aligned one is not provided. 
On the other hand, in studies of spatial heterogeneity, they have the prior knowledge of how the districts are divided into.

\section{Comparisons of Debiasing Performance on Unbiased Test Sets}
\label{supple:comparison-total-table}
Due to the page limit, we intentionally omitted the standard deviation in Table~1 of the main paper. 
Therefore, we present the quantitative results on the unbiased test sets including the standard deviations in Table~\ref{tab:supp_best_test_v1} and Table~\ref{tab:supp_best_test_v2}. 
Note that the results are averaged over five individual trials.

\end{document}